\documentclass{article}
\usepackage[square,numbers,compress]{natbib}
\usepackage{xcolor}
\definecolor{lightgreen}{RGB}{20,210,20}
\usepackage[citecolor=lightgreen]{hyperref}

\PassOptionsToPackage{numbers, compress}{natbib}

\usepackage[final]{neurips_2024}

\usepackage[utf8]{inputenc} 
\usepackage[T1]{fontenc}    
\usepackage{hyperref}       
\hypersetup{
    colorlinks=true,
    linkcolor=blue,
    filecolor=magenta,      
    urlcolor=cyan,
    pdftitle={Overleaf Example},
    pdfpagemode=FullScreen,
    }

\makeatletter
\renewcommand{\@makefnmark}{\textcolor{black}{\@thefnmark}}

\usepackage{makecell}
\usepackage{hhline}
\usepackage{url}            
\usepackage{booktabs}       
\usepackage{amsfonts}       
\usepackage{nicefrac}       
\usepackage{microtype}      
\usepackage{xcolor}         
\usepackage{diagbox}
\usepackage{graphicx}
\usepackage{amsmath}
\usepackage{amssymb}
\usepackage{booktabs}
\usepackage[utf8]{inputenc} 
\usepackage[T1]{fontenc}    
\usepackage{url}            
\usepackage{amsfonts}       
\usepackage{nicefrac}       
\usepackage{microtype}      
\usepackage{xcolor}
\usepackage{times}
\usepackage{bbding}
\usepackage{multirow}
\usepackage{paralist, tabularx}
\usepackage{caption}
\usepackage{subcaption}
\usepackage{pifont}
\usepackage{enumitem}
\usepackage{newtxtext}
\usepackage{wrapfig}
\usepackage{floatrow}
\usepackage{graphicx}
\usepackage{float}
\usepackage{xspace}
\usepackage{natbib}
\usepackage{appendix}
\usepackage{array,etoolbox}

\newfloatcommand{capbtabbox}{table}[][\FBwidth]

\usepackage{amsmath, amssymb}
\usepackage{enumitem}

\usepackage{color}
\usepackage{colortbl}
\usepackage{lipsum}
\definecolor{mygray}{gray}{.9}
\makeatletter
\newcommand{\thickhline}{%
    \noalign {\ifnum 0=`}\fi \hrule height 1pt
    \futurelet \reserved@a \@xhline
}
\usepackage{tikz}
\usetikzlibrary{tikzmark}

\usepackage{arydshln}
\usepackage{adjustbox}

\def\pair#1#2{$\langle \textit{#1}, \textit{#2} \rangle$}
\def\Tideo{{Tideo}}
\def\Tideos{{Tideos}}

\definecolor{darkgreen}{rgb}{0,0.5,0}
\definecolor{azureblue}{rgb}{0,0.5,1}
\definecolor{darkgreen}{rgb}{1,0,0}
\definecolor{color1}{HTML}{006EB8}
\definecolor{color2}{HTML}{009B55}

\usepackage{pifont}

\usepackage[capitalize]{cleveref}
\crefname{section}{Sec.}{Secs.}
\Crefname{section}{Section}{Sections}
\Crefname{table}{Table}{Tables}
\crefname{table}{Tab.}{Tabs.}
\crefname{appendix}{Sec.}{Secs.}
\Crefname{appendix}{Section}{Sections}

\usepackage{color,colortbl}
\definecolor{darkgreen}{rgb}{0,0.5,0}

\definecolor{darkgreen}{rgb}{1,0,0}

\definecolor{azureblue}{rgb}{0,0.5,1}

\newcommand{\sevila}{\textsc{SeViLA}\xspace}

\newcommand{\ie}{\textit{i}.\textit{e}.,}
\newcommand{\eg}{\textit{e}.\textit{g}.,}
\newcommand{\cf}{\textit{c}.\textit{f}.}

\definecolor{navy}{RGB}{0, 80, 200} 
\definecolor{maroon}{RGB}{0, 150, 0} 

\newcommand{\tline}{%
    \noalign {\ifnum 0=`}\fi \hrule height 1pt
    \futurelet \reserved@a \@xhline
}

\title{TOPA: Extending Large Language Models for \\ Video Understanding via Text-Only Pre-Alignment}

\author{
  Wei Li\textsuperscript{\rm 1,\rm 2} \qquad Hehe Fan\textsuperscript{\rm 1}\footnotemark[1]\qquad Yongkang Wong\textsuperscript{\rm 3}\qquad Mohan Kankanhalli\textsuperscript{\rm 3}\qquad Yi Yang\textsuperscript{\rm 1,\rm 2}\\
  \textsuperscript{\rm 1} ReLER Lab, CCAI, Zhejiang University, China\\
  \textsuperscript{\rm 2} The State Key Laboratory of Brain-Machine Intelligence, Zhejiang University, China\\
  \textsuperscript{\rm 3} School of Computing, National University of Singapore, Singapore\\
   \texttt{\{weili6,hehefan,yangyics\}@zju.edu.cn} \\ 
   \texttt{yongkang.wong@nus.edu.sg} \qquad 
   \texttt{mohan@comp.nus.edu.sg} \\
   \texttt{\url{https://github.com/dhg-wei/TOPA}}
}

\begin{document}

\maketitle
\renewcommand{\thefootnote}{\textcolor{black}{*}}
\footnotetext[1]{\textcolor{black}{Hehe Fan is the corresponding author.}}

\begin{abstract}
Recent advancements in image understanding have benefited from the extensive use of web image-text pairs. 
However, video understanding remains a challenge despite the availability of substantial web video-text data. This difficulty primarily arises from the inherent complexity of videos and the inefficient language supervision in recent web-collected video-text datasets.
In this paper, we introduce Text-Only Pre-Alignment~(TOPA), a novel approach to extend large language models (LLMs) for video understanding, without the need for pre-training on real video data.
Specifically, we first employ an advanced LLM to automatically generate \textit{Textual Videos} comprising continuous textual frames, along with corresponding annotations to simulate real video-text pairs. 
Then, these annotated textual videos are used to pre-align language-only LLMs with the video modality. 
To bridge the gap between textual and real videos, we employ the CLIP model as the feature extractor to align image and text modalities.  
During text-only pre-alignment, the continuous textual frames, encoded as a sequence of  CLIP text features, are analogous to continuous CLIP image features, thus aligning the LLM with real video representation. 
Extensive experiments, including zero-shot evaluation and finetuning on various video understanding tasks, demonstrate that TOPA is an effective and efficient framework for aligning video modality with LLMs. 
In particular, without training on any video data, the TOPA-Llama2-13B model achieves a Top-1 accuracy of 51.0\% on the challenging long-form video understanding benchmark, EgoSchema. 
This performance surpasses previous video-text pre-training approaches and is competitive with recent GPT-3.5-based video agents.

\end{abstract}

\section{Introduction}
\vspace{-0.1in}
Image-language understanding has made large advancements in both image-language alignment~\cite{blip1,clip} and Multimodal Large Language Models~(MLLMs)~\cite{alayrac2022flamingo,li2023blip,llava,minigpt}, aided by pre-training on large-scale noise-paired image-text data collected from the web~\cite{cc12m,align,laion400m,cc3m,laion5b}. 
This raises a question: \textit{Can we mirror this success in video-language understanding?} 
Research~\cite{longvivit,wang2022internvideo,frozenbilm,zhao2024videoprism}  has explored pretraining video-language models on millions of web video-text data~\cite{webvid,miech2019howto100m,wang2023internvid}, achieving promising results in basic  video tasks such as video-text retrieval, video captioning, and video question answering across conventional video benchmarks. 
However, recent research reveals that these models struggle with a challenging long-form video understanding benchmark,~\ie~EgoSchema~\cite{EgoSchema}, which requires intrinsic temporal understanding capabilities. 
This highlights the gap in adapting web video-text pretrained models to more comprehensive video understanding tasks.

We attribute this gap to two primary factors:  
\textit{1) The intrinsic complexity of the video modality.} 
Videos introduce intrinsic complexities in both spatial and temporal dimensions, which are not present in static images. These complexities require extensive training on larger-scale data to effectively capture video dynamics. Furthermore, representing videos typically involves processing multiple frames, significantly increasing computational demands compared to image modeling. The dual challenges of large-scale training and increased computational requirements make video-language modeling particularly challenging.
\textit{2) The limitations of web language supervision.}
The language supervision in recent web video-text datasets primarily comes from subtitles or descriptions associated with the videos~\cite{webvid,miech2019howto100m}. 
However, subtitles often suffer from the issues of visual-textual misalignment~\cite{lin2024noise_align,han2022temporal_align}.
Moreover, the form of descriptive supervision is inefficient in building robust video reasoning capabilities, especially in terms of temporal reasoning. 
This mismatch between the complex video content and the limited supervision hinders effective video-language modeling.

In this paper, we propose an innovative approach to develop video understanding capabilities by using LLMs to simulate and understand video dynamics.
Instead of directly aligning LLMs with real video representation, we first introduce a textual video representation — a sequence of textual frames designed to mimic real visual dynamics. 
This textual video can be readily generated by advanced LLMs and effectively simulates various video dynamics by describing them in text.
Specifically, we present a Textual Video~(TextVid) dataset, automatically generated by LLMs.
TextVid includes: 1) \textit{Textual videos}~(hereinafter referred to as {``\textbf{Tideo}''}), which consist of a sequence of textual frames crafted to mimic the keyframes of real videos, and 2) \textit{\Tideo~annotations}, including comprehensive \Tideo-level dense descriptions and varied question-answer (QA) pairs. These annotations are of high quality and closely align with the \Tideo~content, by virtue of the powerful capability of LLM in language generation.

Building on the proposed TextVid dataset, we introduce the Text-Only Pre-Alignment (TOPA) framework, to effectively and efficiently pre-align LLMs with the video modality, reducing the need for costly video-text pre-training.
We introduce three tasks for video-LLM pre-alignment: \Tideo~summarization, \Tideo~QA and multi-choice \Tideo~QA. 
To bridge the gap between textual \Tideos~and visual videos, we leverage the CLIP~\cite{clip} model for feature extraction. 
Specifically, we employ the CLIP text encoder to extract frame-level representations for {\Tideos}, and the CLIP visual encoder for real videos. 
During the text-only pre-alignment phase, the LLM learns to process continuous CLIP text features of \Tideos. 
In the real video inference phase, it transitions to handling continuous CLIP image features of real video. 
Due to the aligned CLIP image-text feature space, the LLM can adapt to real video inputs despite being trained on textual representations. 
Our main contributions include:

(1) We propose a novel Text-Only Pre-Alignment (TOPA) framework to extend Large Language Models (LLMs) for video understanding. TOPA aligns LLMs with the video modality efficiently and effectively without the need for training on real videos, reducing the costs for video-text pre-training.

(2) We introduce TextVid, a textual video dataset automatically generated by advanced LLMs. TextVid dataset comprises 721K diverse {\Tideos} along with associated high-quality annotations, which include detailed {\Tideo} descriptions and a variety of question-answer pairs.

(3) Extensive experiments demonstrate TOPA's effectiveness across various video understanding tasks. Particularly, the TOPA-Llama2-13B model achieves 51.0\% Top-1 accuracy in the challenging EgoSchema benchmark, outperforming previous video-text pretraining methods and competitive with recent GPT-3.5-based video agents. 

\section{Related Work}
\vspace{-0.1cm}
\textbf{Vision-language alignment.}
CLIP~\cite{clip} aligns the vision and language modalities in a common feature space via contrastive learning with large-scale web image-text data.
MLLMs~\cite{alayrac2022flamingo,li2023blip,llava,minigpt} align the visual model with LLM via training on image-caption pairs and interleaved image-text data.
Video-LLMs~\cite{chen2023videollm,LLamavqa,video-llava,videollama} explore modeling video sequences within LLM spaces, leveraging LLM for video-language understanding.
In this paper, we focus on video-LLM alignment. Rather than using multimodal data for vision-language alignment, we introduce a novel text-only pre-alignment framework to extend LLMs for video understanding without pre-training on real video-text data.

\textbf{LLMs for multimodal data augmentation.}
Recent research explores the use of LLMs to enhance the multimodal data. A line of work~\cite{instructpix2pix,cliprewrire,llava} use LLMs for refining captions or extending the image caption pairs to diverse visual tasks like visual conversation and image editing. Another line of work~\cite{li2023videochat,luo2023valley_videoinstruct,videochatgpt,qian2024momentorvideo} further employ advanced LLM to enrich web video supervision for video instruction tuning.
In this paper, rather than enhancing multimodal datasets, we propose generating text-only data consisting of "textual videos" and diverse language supervision, which aims to simulate real videos and their corresponding annotations.

\textbf{Long-form video understanding.}
 Long-form video understanding~\cite{EgoSchema,wang2024lvbench,wu2024longvideobench} presents significant challenges due to the intricate spatial and temporal dynamics. 
 Conventional video-text pretraining approaches~\cite{MC:MemoryConsolidation,longvivit,wang2023internvid,internvideo2,frozenbilm,zhu2020actbert} utilize extensive web video-caption data for video-language alignment. 
 Recent research~\cite{li2023videochat,internvideo2,videollama,zhao2024distilling} employ video instruction-tuning for video-LLM alignment to enhance video-language understanding.
Another line of research \cite{imagegrid,song2023moviechat,sevila,onepass} seeks to adapt recent image MLLMs to video understanding.
 A parallel line of research~\cite{agent_proviq,morevqa,vipergpt,wang2023vamos,agent_video,videoagent2,agent_dlam,LLovi,wang2024lifelongmemory,langrepo} combine the LLM with various VLM tools as video agents to perform video-understanding tasks.
 In this paper, we propose a novel text-only pre-alignment framework to efficiently and effectively align LLMs with videos without pre-training on real videos.

\begin{figure*}[!t]
    \centering
    \includegraphics[width=1.0\textwidth]{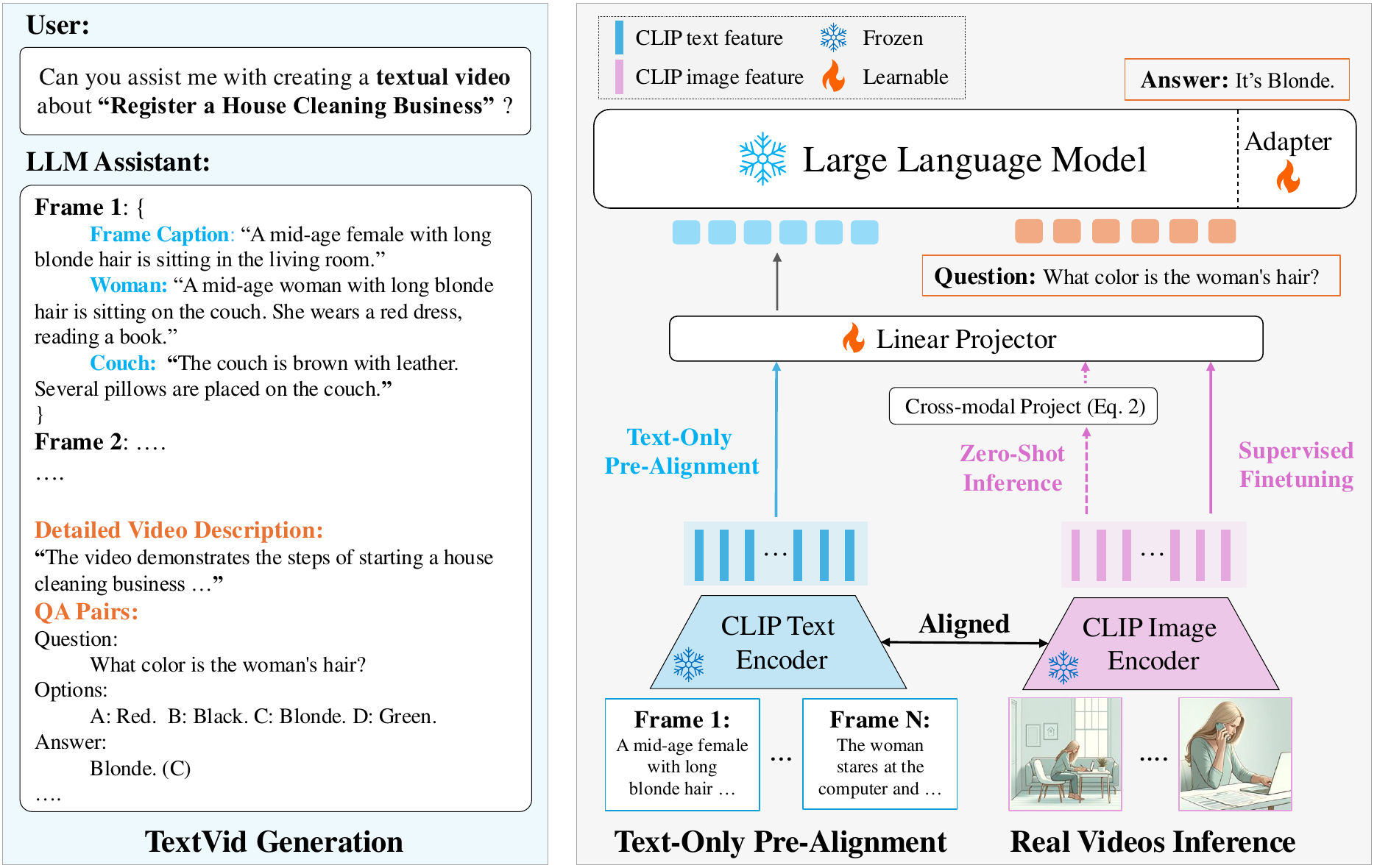}
    \caption{Overview of the proposed Text-Only Pre-Alignment (TOPA) framework. \textit{Left}: The pipeline used for generating the TextVid dataset.  \textit{Right}: The video-LLM alignment framework. 
    During text-only pre-alignment, the LLM learns to process continuous CLIP text features. 
    In zero-shot inference, the LLM uses projected CLIP visual features as input. 
    Additionally, TOPA supports supervised fine-tuning on downstream video datasets to further improve the performance.}
    \label{fig:framework}
    \vspace{-0.1in}
\end{figure*}

\section{Method}

In this section, we detail the TOPA framework. 
We first introduce the data generation pipeline of TextVid Dataset~(Section~\ref{method:textualvideo}). Next, we describe how to align the {\Tideo} representation with LLM~(Section~\ref{Sec:Method-pretrain}). Finally, we discuss adapting the text-only aligned video-LLM model for real video inference~(Section~\ref{method:zero-shot inference}). An overview is illustrated in Figure~\ref{fig:framework}.

\subsection{TextVid Dataset} 
\label{method:textualvideo}

This dataset, comprising textual videos~(\Tideos) and associated annotations, is generated by an advanced LLM (\ie~Gemini Pro 1.0~\cite{gemini}). 
The data generation pipeline is detailed in Appendix~\ref{app:dataset}.
Each {\Tideo} is presented in a textual format and contains 5-15 sequential frames. Each frame includes a frame caption that describes the scene and multiple object captions. To enhance understanding and interaction with these \Tideos, the dataset features a dense description summarizing the \Tideo, as well as a set of multiple-choice questions and answers related to the {\Tideo} content.
The structure of each element is as follows:

\begin{center}
\colorbox{gray!10}{%
\begin{minipage}{\textwidth}
    \begin{tabbing}
    \hspace*{2em}\= \hspace*{2em}\= \hspace*{2em}\= \kill
    \textbf{Dataset Element}: \\
    \> \textbf{\Tideo}: Sequence of textual frames $\{T_1, T_2, \ldots, T_n\}, 5 \leq n \leq 15$ \\
    \>\> For each frame $T_i$: \\
    \>\>\> Frame caption: $C_i$ \\
    \>\>\> Object captions: $D_{i,j}$ for main objects in $T_i$\\
    \> \textbf{Annotations}: \\
    \>\> Global Dense Description of the \Tideo: $D_V$ \\
    \>\> Set of Questions-Options-Answers: $\{(Q_k, O_k, A_k)\}$
    \end{tabbing}
\end{minipage}
}
\end{center}

There are two major advantages of the TextVid dataset. 
\textbf{(1) The large-scale and diverse Tideos.} As the dataset is text-only and fully generated by an LLM, the size of TextVid is practically unlimited. 
Moreover, the {\Tideos} can cover a broad range of domains by simply prompting the language model with appropriate conditions. It is distinctly different from previous web video-text dataset like Howto100M~\cite{miech2019howto100m} that are limited to specific human-centric instructional videos. In practice, we enhance the diversity of TextVid by randomly sampling video captions from WebVid-2M~\cite{webvid}, video titles from Howto100m~\cite{miech2019howto100m}, video tasks from Ego4D~\cite{ego4d} and object names with descriptions from WordNet~\cite{miller1995wordnet} as a condition of prompts. These varied prompts enable the language model to generate a diverse dataset.
\textbf{(2) The high-quality, consistent and free-form language supervision.}
The language supervisions are generated along with {\Tideos}. The advanced capabilities of LLM ensure the quality of these supervisions, making them less noisy than web video-text data. Moreover, both the {\Tideo} and the supervision are in textual format, making the supervision closely aligned with the {\Tideo}'s content. Additionally, the format of the language supervision is unrestricted. For example, we prompt the LLM to generate dense descriptions and multi-choice QA pairs as language supervision.

\subsection{Text-Only Pre-Alignment}  \label{Sec:Method-pretrain}
\textbf{Preliminary: Video-LLM alignment.} The goal of video-LLM alignment is to extend pre-trained LLMs for processing video inputs.
Given a video sampled with $n$ frames $\{\mathbf{I}_1, \mathbf{I}_2, \ldots, \mathbf{I}_n\}$,
Recent work~\cite{LLamavqa,frozenbilm} uses a frozen CLIP model to extract the frame-level visual feature, formulated as $\mathbf{f}^v_i = {E}_{\text{image}}(\mathbf{I}_i)$, where $E_{\text{image}}$ denotes CLIP image encoder. The CLIP features are then projected into the LLM space via a simple linear layer, denoted as ${G}({P}(\mathbf{f}^v_1),...,{P}(\mathbf{f}^v_n))$, where ${G}$ denotes a language model and ${P}$ denotes a projection layer that projects the CLIP feature to LLM space.

\textbf{Tideo representation.} In this work, we leverage {\Tideos} (\cf~Section~\ref{method:textualvideo}) for video-LLM pre-alignment instead of training on real videos. Specifically, given the textual frame $T_i$, we employ CLIP text encoder to extract the frame representation from frame caption $C_i$ and detailed object captions $D_i$, represented as $\mathbf{f}^t_i = {F}_{\text{fusion}}({E}_{\text{text}}(C_i),{E}_{\text{text}}(D_{i,1}),...,{E}_{\text{text}}(D_{i,j}))$, where $F_{\text{fusion}}$ is a fusion function such as simple average pooling, and $E_{\text{text}}$ denotes the CLIP text encoder. 
A {\Tideo} with $n$ textual frames is represented as $\mathbf{V}^t=\{\mathbf{f}^t_1,...,\mathbf{f}^t_n\}$.

\textbf{Text-only pre-alignment.}
Given the {\Tideo} $T$, dense {\Tideo}-level description $D_V$, and QA pairs with multiple choices $\{(Q_k,O_k,A_k)\}$, we introduce the following tasks for {\Tideo}-LLM alignment:
(1)~\textbf{\Tideo~Summarization}: Given the {\Tideo}, generate a detailed description to summarize the {\Tideo}; 
(2)~\textbf{\Tideo~QA}: Given the {\Tideo} and question, predict the answer; 
(3)~\textbf{Multi-choice \Tideo~QA}: Given the {\Tideo}, question and multiple choices, choose the correct answer from the candidates.
We employ a unified auto-regressive Language Modeling~(LM) objective for these three tasks:
\begin{equation}
    \mathcal{L}_{\text{LM}}(\theta_{{G}},\theta_{P})=-\frac{1}{| t |} \sum_{i=1}^{|t|}\log {G}\big(t_i|{P}(\mathbf{V^t}),{Z},t_{<i} \big),
\end{equation}
where $\mathbf{V^t}$ denotes the {\Tideo} representation, and $\mathbf{V^t}=\{\mathbf{f}^t_1,...,\mathbf{f}^t_n\}$ during the text-only training, Z denotes the task specific condition tokens and $t_i$ denotes the $i_{th}$ target token. $\theta_{{G}}$ and $\theta_{{P}}$ denote the learnable parameters of the LLM adapter and the projection layer $P$, respectively.
In practice, we use the following format as the LLM input:
\textcolor{navy}{\{Task Instruction\}.~Video:\{$\mathbf{f}^t_1$, ..., $\mathbf{f}^t_n$\}. \{Task Conditions\}. Answer:} \textcolor{maroon}{\{Predict Targets\}}.
For the \Tideo~summarization task, the target is detailed Tideo descriptions. For {\Tideo} QA task, the target is the answer and the condition is the question. For multi-choice {\Tideo} QA task, the target is the correct option and the condition consists with question and options. 
The details of the task-specific prompts are included in Appendix~\ref{app:text-only task prompts}.

\subsection{Adapting to Real Video Understanding} \label{method:zero-shot inference}
Section~\ref{Sec:Method-pretrain} introduces the text-only pre-alignment using the TextVid dataset. In this section, we detail how to adapt this text-only pre-aligned LLM for real video understanding. We introduce two approaches: one is zero-shot inference, which directly infers with real video data. And the other is supervised finetuning, where the pre-aligned model is further finetuned on downstream video data.

\textbf{Zero-shot inference.} 
During pre-alignment, we leverage the textual representation $\mathbf{V}^t=\{\mathbf{f}^t_1,...,\mathbf{f}^t_n\}$ as the Tideo representation. During inference, we take real videos features as input, \ie~$\mathbf{V}^v=\{\mathbf{f}^v_1,...,\mathbf{f}^v_n\}$, where $\mathbf{f}^v_i = E_{\text{image}}(\mathbf{I}_i)$. These two modality features $\mathbf{f}^t$ and $\mathbf{f}^v$ that come from CLIP image encoder and CLIP text encoder are aligned via CLIP pre-training. This aligned image-text representation makes it possible to perform zero-shot inference without additional finetuning. However, the \textit{modality gap} phenomenon~\cite{gapnoimages,li2023decap,mindthegap,gapcapdec,gapC3}, \ie~CLIP image feature and CLIP text feature are located in two completely separate regions of the feature space, prevents us from directly taking the visual feature $\mathbf{f}^v$ as the textual feature $\mathbf{f}^t$.
To bridge this modality gap, we follow DeCap~\cite{li2023decap} to employ a support memory to project the CLIP visual feature into the CLIP text feature space. This training-free projection process is formulated as:
\begin{equation}
\label{eq:project}
\mathbf{f}^{v\rightarrow t} = \sum_{i=1}^N w_i* \mathbf{m}_i = \sum_{i=1}^N
\frac{\exp((\mathbf{m}_i^{\top} \mathbf{f}^v) / \tau)}{\sum_{k=1}^{N}\exp((\mathbf{m}_k^{\top} \mathbf{f}^v)/ \tau)}* \mathbf{m}_i,
\end{equation}
where $\mathbf{m}_i$ denotes CLIP text feature from a pre-constructed memory of size $N$, $\mathbf{f}^v$ denotes input frame feature of real video and $\mathbf{f}^{v\rightarrow t}$ denotes the projected feature. During zero-shot inference, we take the $\mathbf{V}^{{v\rightarrow t}}= \{\mathbf{f}^{v\rightarrow t}_1, ..., \mathbf{f}^{{v\rightarrow t}}_n\}$ as the real video's representation.

\textbf{Supervised finetuning.} On the other hand, the text-only pre-alignment can be viewed as a pretraining stage. Following the pretraining-finetuning paradigm, the pre-aligned LLMs can then be fine-tuned on real video data for improved downstream task performance. The finetuning process is similar to the text-only pre-alignment as detailed in Section~\ref{Sec:Method-pretrain}, except that the LLM receives a sequence of CLIP visual features as input instead of CLIP textual features.

\subsection{Implementation Details} 
\label{exp_details}

We leverage Llama2-7B, Llama2-13B~\cite{llama2} and Llama3-8B as the LLM backbone. Additionally, we employ the Llama-adapter~\cite{zhang2023llamaadapter} with an adaptation embedding length of 50.
We utilize CLIP-ViT-L as the multimodal encoder. We employ a simple linear layer to project the CLIP feature into the LLM feature space.
During training, the CLIP model and LLM backbone are frozen. The projection layer and additional Llama-adapter are trainable.
For text-only pre-alignment, 
we uniformly sample the {\Tideos} into 10 frames. 
We train the model on a mixture of tasks comprising {\Tideo} summarization, {\Tideo} QA, multi-choice {\Tideo} QA with the ratio of 1:1:2. 
For zero-shot inference, we construct a memory for cross-modal projection, consisting of 2M CLIP text features sampled from captions in the TextVid dataset.
TOPA-Llama2-7B and TOPA-Llama3-8B are trained on four 40G-A100 GPUs in one day. TOPA-Llama2-13B is trained in two days.
More training details of TOPA and baselines are included in Appendix~\ref{app:training_details}.

\section{Experiments}
\begin{figure*}[t]
    \centering
    \includegraphics[width=0.9\textwidth]{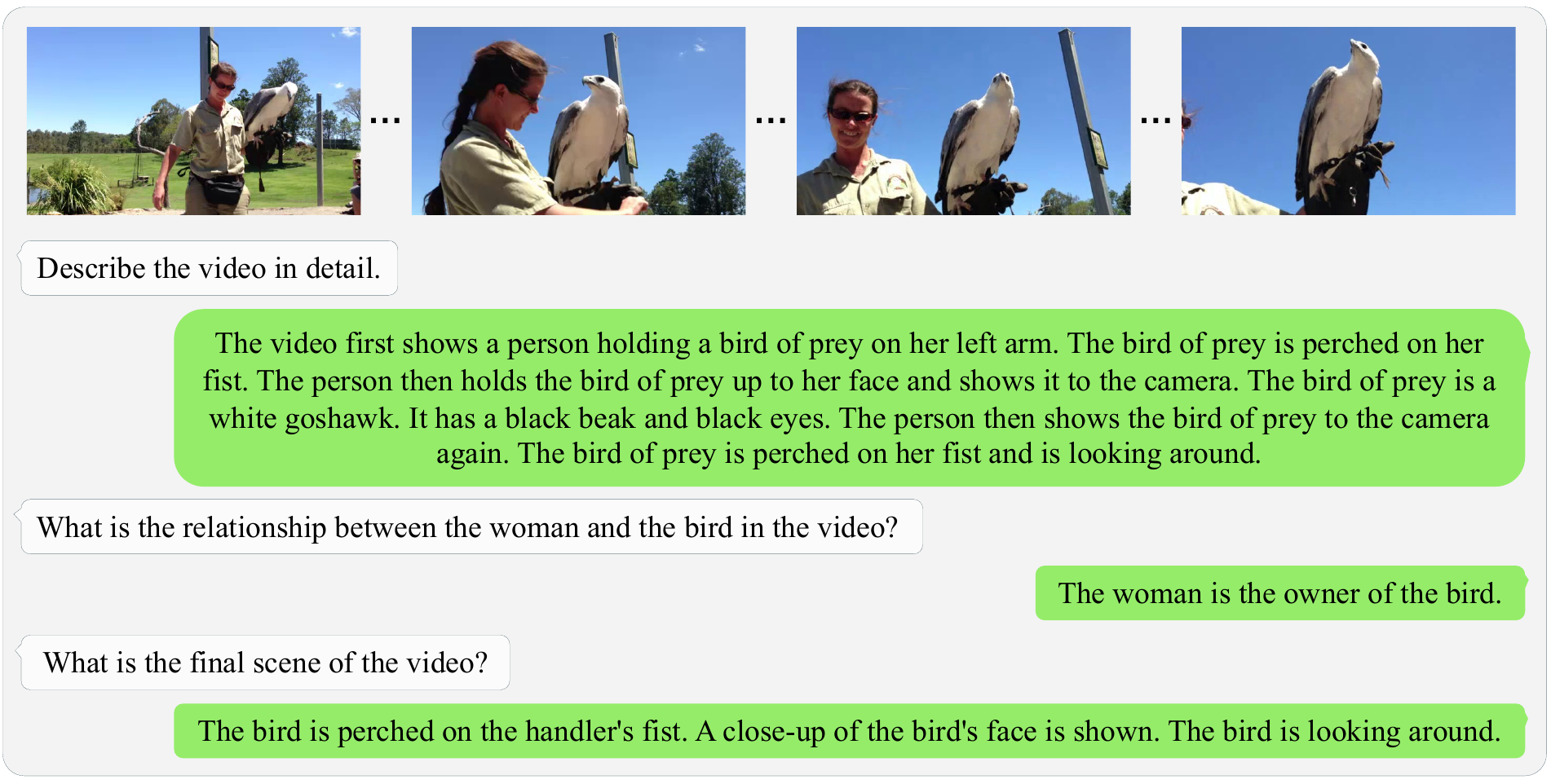}
    \vspace{-0.5em}
    \caption{Examples of TOPA-LLama2-13B for video-language understanding. Given a video, TOPA is able to summarize the video content and answer the questions.}
    \label{fig:fig2}
\end{figure*}
\vspace{-0.2cm}
TOPA enables the LLM to perform various video understanding tasks as shown in Figure~\ref{fig:fig2}.
In this section, we evaluate TOPA on multi-choice video QA and video captioning tasks.
Section~\ref{exp:zero-shot multi-choice QA} evaluates TOPA on NeXT-QA~\cite{nextqa}, STAR~\cite{star}, TVQA~\cite{tvqa}, recent challenging EgoSchema~\cite{EgoSchema} and MVBench\cite{li2023mvbench} benchmarks with the zero-shot setting.
We further evaluate TOPA on multi-choice video QA with the finetuning setting (Section~\ref{exp:finetuning}) and zero-shot video captioning task (Section~\ref{exp:video captioning}). 
In Section~\ref{exp:ablations}, we conduct ablation study on the LLM prior and input video frames.
We report Top-1 accuracy on multi-choice video QA benchmarks and CIDEr~\cite{vedantam2015cider} score on video captioning benchmarks.
We mainly compare TOPA with the following categories of video understanding approaches:

\textbf{(1) Web video pre-training approaches}~\citep{MC:MemoryConsolidation,longvivit,wang2022internvideo,internvideo2,frozenbilm}.
This line of work aims to develop general video-language models by leveraging extensive web videos, using associated video captions or audio as weak supervision signals.

\textbf{(2) Adapting image MLLMs for video understanding}~\citep{imagegrid,sevila,onepass}.
These approaches aim to extend the image understanding capabilities of recent vision-language models (VLMs) to video understanding.
Specifically, SeViLa~\cite{sevila} utilizes BLIP-2 for localizing and understanding key frames of a video. 
IG-VLM~\cite{imagegrid} converts video into a composite image by arranging the video frames into a grid layout.

\textbf{(3) LLM-based video agents}~\citep{morevqa,agent_chatvideo,wang2023vamos,agent_video,LLovi,videoagent2,langrepo,wang2024lifelongmemory}. 
This line of work leverages LLMs like GPT-3.5 and GPT-4 as an agent to understand a video by designing and executing a series of actions. 
The language-only agents perceive visual information via recent foundation VLMs (\eg~CLIP~\cite{clip}, BLIP-2~\cite{li2023blip}, LaViLa~\cite{lavila} and PALI~\cite{chen2022pali}).

\textbf{(4) Our text-only pre-alignment.} Different from the above works, TOPA leverages the proposed TextVid dataset for video-LLM pre-alignment, enabling the LLM to process continuous features. Thus, it can enable performing video understanding tasks.

\vspace{-0.1cm}
\subsection{Zero-Shot Evaluation on Multi-Choice Video QA}\label{exp:zero-shot multi-choice QA}
\vspace{-0.1cm}

\subsubsection{Zero-shot Results on EgoSchema} \label{exp:EgoSchema}

\begin{table}[ht] 
\small
\vspace{-1em}
\caption{Zero-shot results on EgoSchema~\cite{EgoSchema} full set. Methods that
leverage closed-source LLMs are marked in gray. $\dagger$ denotes the model is trained with in-domain egocentric videos from Ego4D~\cite{ego4d}. \\$^*$ denotes results on EgoSchema subset. Results of InternVideo and FrozenBiLM are sourced from ~\cite{EgoSchema}. Results of \sevila{} are sourced from \cite{longvivit}.
}
\label{tab:egos}
\centering
\setlength{\tabcolsep}{3mm}
\begin{tabular}{rrllc}
\tline
&&\textbf{Core VLMs}&\textbf{Core LLMs}&\textbf{Acc@1}\\
\hline

\multicolumn{2}{r}{
Human Eval \cite{EgoSchema}\textcolor{gray}{[NeurIPS\space2023]}}&-&-&75.0\\
\rowcolor{gray!20}\multicolumn{2}{r}{Gemini-1.5-Pro~\cite{gemini1.5}\textcolor{gray}{[arXiv 2024.2\space\space]}}&-&Gemini-1.5-Pro&63.2\\

\hline

 \multicolumn{2}{l}{\textit{(Pre-train on web video-text data)}}   \\
 \multicolumn{2}{r}{FrozenBiLM~\cite{frozenbilm}\textcolor{gray}{[NeurIPS\space2022]}}&-&-&26.9 \\
\multicolumn{2}{r}{InternVideo~\cite{wang2022internvideo}\textcolor{gray}{[arXiv 2022.12]}}&-&-&32.1\\
\multicolumn{2}{r}{LongViViT~\cite{longvivit}\textcolor{gray}{[CVPR~~2024\space\space\space]}}&-&-& 33.3\\
\multicolumn{2}{r}{MC-ViT-L$^\dagger$~\cite{MC:MemoryConsolidation}\textcolor{gray}{[ICML ~2024\space\space\space]}}&-&-&44.4\\
\multicolumn{2}{r}{InternVideo2$_{s3}$-6B$^\dagger$~\cite{internvideo2}\textcolor{gray}{[arXiv 2024.3\space\space]}}&-&-&41.1 \\
\hline
\multicolumn{2}{l}{\textit{(Adapt image MLLMs for video understanding)}}   \\
\multicolumn{2}{r}{\sevila{} \cite{sevila}\textcolor{gray}{[NeurIPS\space2023]}}&BLIP-2&FLAN-T5-XL~\cite{T5}& 22.7\\

\multicolumn{2}{r}{MVU \cite{onepass}\textcolor{gray}{[arXiv 2024.3\space\space]}} &LLaVA-v1.5-13B& Vicuna-13B&37.6 \\

\multicolumn{2}{r}{IG-VLM \cite{imagegrid}\textcolor{gray}{[arXiv 2024.3\space\space]} }&LLaVA v1.6-7B&Vicuna-7B& 35.8$^*$\\
\multicolumn{2}{r}{IG-VLM \cite{imagegrid}\textcolor{gray}{[arXiv 2024.3\space\space]}} &LLaVA v1.6-13B& Vicuna-13B&47.0$^*$ \\

\hline
\multicolumn{2}{l}{\textit{(LLM-based video agents)}}   \\
\multicolumn{2}{r}{LangRepo~\cite{langrepo}\textcolor{gray}{[arXiv 2024.3\space\space]}}&CLIP-ViT-L& Mixtral-12B~\cite{jiang2024mixtral}&41.2\\

\multicolumn{2}{r}{Vamos~\cite{wang2023vamos}\textcolor{gray}{[arXiv 2023.11]}}&BLIP-2&Llama2-13B& 36.7$^*$\\

\rowcolor{gray!20}\multicolumn{2}{r}{{Vamos}~\cite{wang2023vamos}\textcolor{gray}{[arXiv 2023.11]}}&BLIP-2&GPT-3.5& 41.2$^*$\\

\rowcolor{gray!20}\multicolumn{2}{r}{{Vamos}~\cite{wang2023vamos}\textcolor{gray}{[arXiv 2023.11]}}&BLIP-2&GPT-4& 48.3$^*$\\

\rowcolor{gray!20}\multicolumn{2}{r}{MoReVQA~\cite{morevqa}\textcolor{gray}{[CVPR 2024 \space\space\space]}}&PALI-3-5B~\cite{chen2022pali} & PaLM-2~\cite{anil2023palm}&51.7\\

\rowcolor{gray!20}\multicolumn{2}{r}{LLoVi~\cite{LLovi}\textcolor{gray}{[arXiv 2024.3\space\space]}}&LaViLa$^\dagger$&GPT-3.5&50.3\\

\rowcolor{gray!20}\multicolumn{2}{r}{VideoAgent~\cite{agent_video}\textcolor{gray}{[ECCV 2024 \space\space\space]}}&LaViLa$^\dagger$&GPT-4&54.1\\

\rowcolor{gray!20}\multicolumn{2}{r}{LifelongMemory~\cite{wang2024lifelongmemory}\textcolor{gray}{[arXiv 2024.3\space\space]}}&LaViLa$^\dagger$ & GPT-4&62.4\\

\rowcolor{gray!20}\multicolumn{2}{r}{VideoAgent~\cite{videoagent2}\textcolor{gray}{[ECCV 2024 \space\space\space]}}&Video-LLava~\cite{video-llava}&GPT-4&60.2\\

\hline


\multirow{3}{*}{\textit{(Our Text-Only Pre-Alignment)}}&\multicolumn{1}{c}{TOPA}&CLIP-ViT-L&Llama2-7B&41.2\\
&\multicolumn{1}{c}{TOPA}&CLIP-ViT-L&Llama3-8B&44.2\\
&\multicolumn{1}{c}{TOPA}&CLIP-ViT-L&Llama2-13B&51.0\\

\tline
\vspace{-1em}
\end{tabular}
\end{table}

Table \ref{tab:egos} shows the results on EgoSchema full set. We compare our method against a range of recent approaches in video understanding. Our proposed text-only pre-alignment framework, despite training without real videos, shows impressive results on the EgoSchema benchmark. TOPA outperforms previous image-based adaptation approach IG-VLM and video agents LLoVi and Vamos with the same scale LLM~(Llama2-7B and Llama2-13B). Moreover, TOPA shows  consistent improvements when scaled up with a larger LLM backbone, indicating the effectiveness of LLMs in complex video understanding tasks.

\textbf{Discussion 1: The necessity of high-quality language supervision for video understanding.}
Recent video pre-training approaches like LongViVit~\cite{longvivit} and InternVideo~\cite{wang2022internvideo}, despite training on million-level web video-text data, show inferior performance on EgoSchema evaluation. These results highlight the inefficacy and inefficiency of conventional contrastive pre-training in understanding long-form videos, primarily due to noisy and simplistic language supervision.
In contrast, our TOPA, trained on 721K \Tideos with high-quality language supervision, shows impressive results on EgoSchema. 
It indicates that, unlike image understanding which significantly benefits from leveraging web language as supervision, video understanding may require more precise and accurate language supervision to better capture the complex visual dynamics.

\textbf{Discussion 2: Video agents versus end-to-end video-LLM modeling.}
Video agents have shown impressive results on the EgoSchema benchmark, aided by advanced LLMs and VLMs. However, a significant limitation of these approaches is their heavy reliance on the powerful LLMs. For example, the accuracy of Vamos drops by -11.6\% when the GPT-4 is replaced with Llama2-13B, largely falling behind the performance of the TOPA-Llama2-13B model.
The reliance on powerful closed-source LLMs restricts its application fields and introduces external overheads.
Moreover, video agents make decisions based on the language format clues collected by VLMs. Converting the video content into language clues may lead to a limited upper bound compared to end-to-end modeling.
Additionally, the inference speed of these approaches is another concern, since it involves multiple interactions with both VLMs and LLMs. In contrast, end-to-end video-LLM models, which condense the video into a sequence of embeddings as the input of LLM, are more efficient.

\subsubsection{Zero-shot Results on NExT-QA, STAR and TVQA} \label{exp:mcqa}
Table~\ref{tab:zeroshot} shows the multi-choice video QA results across various benchmarks.
TOPA achieves impressive performance on the TVQA and EgoSchema benchmarks, significantly outperforming previous video pre-training models and image-to-video adaptation approaches. This indicates that our TOPA framework effectively enables LLMs to handle video input, despite not being pre-trained on real videos.
However, for the NeXT-QA and STAR benchmarks, TOPA underperforms compared to SeViLA and IG-VLM. A major reason is that these benchmarks involve many fine-grained visual questions, including those about object locations and relationships. SeViLA and IG-VLM, benefiting from the advanced image-understanding capabilities of pre-trained VLMs such as LLaVA, excel in answering these fine-grained visual questions. 
In contrast, our TOPA framework primarily focuses on high-level semantic alignment. Moreover, during zero-shot inference, we project the visual features into the text feature space to bridge the modality gap, as described in Eq.~\ref{eq:project}. This cross-modal semantic projection process tends to overlook fine-grained visual details, such as object locations, which leads to inferior performance on the STAR benchmark. We provide extensive qualitative results to illustrate TOPA's advantages and limitations across various video understanding tasks in Appendix~\ref{app:Qualitative Results}.

\begin{table}[ht]
\small
\caption{Zero-shot results on multi-choice video QA benchmarks.}
\label{tab:zeroshot}
\vspace{-1em}
\centering
    \setlength{\tabcolsep}{1.mm}
    \begin{tabular}{l cccccccccccc}
    \tline
    \multirow{2}{*}{\textbf{Model (\# Frames)}} & \multicolumn{4}{c}{\textbf{NExT-QA}}& \multicolumn{5}{c}{\textbf{STAR}} & \multirow{2}{*}{\textbf{~~TVQA~~}} & \multirow{2}{*}{\textbf{EgoSchema}} \\
    \cline{2-5}
    \cline{6-10}
    
    & {Tem.} & {Cau.} & {Des.} & \textbf{Avg.} & {Int.} & {Seq.} & {Pre.} & {Fea.} & \textbf{Avg.} &&\\
    \hline
    
    FrozenBiLM (10) \cite{frozenbilm}  & - & - & - & - & - & - & - & - & - & 29.7 & 26.9 \\
    InternVideo (8) \cite{wang2022internvideo} & 43.4 & 48.0 & 65.1 & 49.1 & 43.8 & 
    {43.2} & {42.3} & 37.4 & 41.6  & 35.9 & 32.1 \\
    
    \midrule
    \sevila{}~(32 $\rightarrow$ 4)~\cite{sevila}& {61.3} & {61.5} & {75.6} & {63.6} & {48.3} & {45.0} & {44.4}  & {40.8} & {44.6} & {38.2} & {22.7} \\ 
    IG-VLM-Llava7B (6)~\cite{imagegrid}  &63.1& 57.3& 74.9 &63.1&49.3 &50.1 &48.4& 48.8& 49.6&42.1&35.8\\
    IG-VLM-Llava13B (6)~\cite{imagegrid}~ &61.6 &55.7& 70.8& 61.2&51.5& 52.0& 51.0& 51.8& 51.7&44.5&47.0\\
    \midrule
    TOPA-Llama2-7B~(10)  & 53.4 & 61.3 & 68.3 & 59.9 & 36.4 & 45.6 & 39.3 & 36.3 & 41.3 & 48.2 & 41.2 \\
    TOPA-Llama3-8B~(10)  & 53.0 & 61.9 & 64.5 & 59.5 & 40.8 & 43.1 & 39.4 & 34.5 & 41.4 & 48.5 & 44.2 \\
    TOPA-Llama2-13B~(10) & 57.2 & 63.6 & 68.9 & 62.1 & 41.6 & 46.2 & 44.2 & 36.7 & 43.0 & 50.2 & 51.0 \\
    \tline
    \end{tabular}
\end{table}

\subsubsection{Results on MVBench} \label{MVbench results}
MVBench~\cite{li2023mvbench} is a recent video-language understanding benchmark that covers 20 challenging video tasks, regrouped from existing video-language benchmarks.
Table~\ref{tab:mvpbench} shows the results.
TOPA demonstrates impressive results compared to previous image MLLM and video MLLM. It excels particularly in tasks requiring high-level video-language understanding, such as Scene Transition (ST), Episodic Reasoning (ER), and Unexpected Action (UA).
TOPA Surprisingly excels in the Action Localization (AL) task, which requires identifying the moment an action occurs. This indicates that the text-only pre-alignment enables the LLM to understand temporal visual sequences.
However, TOPA struggles with tasks that demand fine-grained visual understanding, such as Moving Direction (MR), Action Antonym (AA), and Object Shuffle (OS). A common challenge in these tasks is the requirement for detailed visual understanding. For example, Action Antonym involves identifying the direction of an action, while Object Shuffle involves locating objects.
TOPA struggles in  these fine-grained visual tasks since it is trained with CLIP text features. The modality gap between CLIP text features and image features hinders TOPA from capturing visual details. 
Further video instruction tuning might address this limitation, which we leave for future work. We provide qualitative results in Appendix~\ref{app:Qualitative Results} to illustrate TOPA's advantages and limitations on various video understanding tasks.

\begin{table}[ht]
\vspace{-0.2cm}
    \small
    \centering
    \setlength\tabcolsep{2pt}
    \resizebox{1.0\textwidth}{!}{
        \begin{tabular}{l|l|c|c|c|c|c|c|c|c|c|c|c|c|c|c|c|c|c|c|c|c|c}

        \hline
        \textbf{Model} & \textbf{LLM} & \textbf{Avg} & \textbf{AS} & \textbf{AP} & \textbf{AA} & \textbf{FA} & \textbf{UA} & \textbf{OE} & \textbf{OI} & \textbf{OS} & \textbf{MD} & \textbf{AL} & \textbf{ST} & \textbf{AC} & \textbf{MC} & \textbf{MA} & \textbf{SC} & \textbf{FP} & \textbf{CO} & \textbf{EN} & \textbf{ER} & \textbf{CI} \\
        \hline

      Random & - & \cellcolor{gray!20}{27.3} & 25.0 & 25.0 & 33.3 & 25.0 & 25.0 & 33.3 & 25.0 & 33.3 & 25.0 & 25.0 & 25.0 & 33.3 & 25.0 & 33.3 & 33.3 & 25.0 & 33.3 & 25.0 & 20.0 & 30.9 \\
      \hline
 \multicolumn{22}{l}{\textit{\textbf{Image MLLMs}: Following ~\cite{dai2024instructblip}, all models take 4 frames as input, with the output embeddings concatenated before feeding into the LLM.}}   \\

        mPLUG-Owl-I~\cite{xu2023mplug} & LLaMA-7B & \cellcolor{gray!20}{29.4} & 25.0 & 20.0 & 44.5 & 27.0 & 23.5 & 36.0 & 24.0 & 34.0 & 23.0 & 24.0 & 34.5 & 34.5 & 22.0 & 31.5 & 40.0 & 24.0 & 37.0 & 25.5 & 21.0 & 37.0 \\
    BLIP2~\cite{li2023blip} & FlanT5-XL & \cellcolor{gray!20}{31.4} & 24.5 & 29.0 & 33.5 & 17.0 & 42.0 & 51.5 & 26.0 & 31.0 & 25.5 & 26.0 & 32.5 & 25.5 & 30.0 & 40.0 & 42.0 & 27.0 & 30.0 & 26.0 & 37.0 & 31.0 \\
        LLaMA-Adapter~\cite{zhang2023llamaadapter} & LLaMA-7B & \cellcolor{gray!20}{31.7} & 23.0 & 28.0 & 51.0 & 30.0 & 33.0 & 53.5 & 32.5 & 33.5 & 25.5 & 21.5 & 30.5 & 29.0 & 22.5 & 41.5 & 39.5 & 25.0 & 31.5 & 22.5 & 28.0 & 32.0 \\
        Otter-I~\cite{otter} & MPT-7B & \cellcolor{gray!20}{33.5} & 34.5 & 32.0 & 39.5 & 30.5 & 38.5 & 48.5 & 44.0 & 29.5 & 19.0 & 25.5 & 55.0 & 20.0 & 32.5 & 28.5 & 39.0 & 28.0 & 27.0 & 32.0 & 29.0 & 36.5 \\
        MiniGPT-4~\cite{minigpt} & Vicuna-7B & \cellcolor{gray!20}{18.8} & 16.0 & 18.0 & 26.0 & 21.5 & 16.0 & 29.5 & 25.5 & 13.0 & 11.5 & 12.0 & 9.5 & 32.5 & 15.5 & 8.0 & 34.0 & 26.0 & 29.5 & 19.0 & 9.9 & 3.0 \\
        InstructBLIP~\cite{dai2024instructblip} & Vicuna-7B & \cellcolor{gray!20}{32.5} & 20.0 & 16.5 & 46.0 & 24.5 & 46.0 & 51.0 & 26.0 & 37.5 & 22.0 & 23.0 & 46.5 & \textbf{42.5} & 26.5 & 40.5 & 32.0 & 25.5 & 30.0 & 25.5 & 30.5 & 38.0 \\
        LLaVA~\cite{llava} & Vicuna-7B & \cellcolor{gray!20}{36.0} & 28.0 & 39.5 & \textbf{63.0} & 30.5 & 39.0 & 53.0 & 41.0 & \textbf{41.5} & 23.0 & 20.5 & 45.0 & 34.0 & 20.5 & 38.5 & 47.0 & 25.0 & 36.0 & 27.0 & 26.5 & \textbf{42.0} \\
        \hline
 \multicolumn{22}{l}{\textit{\textbf{Video MLLMs}: All models take {16} frames as input}}   \\
        Otter-V~\cite{otter} & LLaMA-7B & \cellcolor{gray!20}{26.8} & 23.0 & 23.0 & 27.5 & 27.0 & 29.5 & 53.0 & 28.0 & 33.0 & 24.5 & 23.5 & 27.5 & 26.0 & \textbf{28.5} & 18.0 & 38.5 & 22.0 & 22.0 & 23.5 & 19.0 & 19.5 \\
        mPLUG-Owl-V~\cite{ye2023mplug-owl} & LLaMA-7B & \cellcolor{gray!20}{29.7} & 22.0 & 28.0 & 34.0 & 29.0 & 29.0 & 40.5 & 27.0 & 31.5 & \textbf{27.0} & 23.0 & 29.0 & 31.5 & 27.0 & 40.0 & 44.0 & 24.0 & 31.0 & 26.0 & 20.5 & 29.5 \\

        \hline
 \multicolumn{22}{l}{\textit{\textbf{Instructed Video MLLMs:}~All models take {16} frames as input, with the exception of VideoChatGPT, which uses {100} frames.}}   \\
         VideoChatGPT~\cite{videochatgpt} & Vicuna-7B & \cellcolor{gray!20}{32.7} & 23.5 & 26.0 & 62.0 & 22.5 & 26.5 & \textbf{54.0} & 28.0 & 40.0 & 23.0 & 20.0 & 31.0 & 30.5 & 25.5 & 39.5 & \textbf{48.5} & 29.0 & 33.0 & 29.5 & 26.0 & 35.5 \\
        VideoLLaMA~\cite{videollama} & Vicuna-7B & \cellcolor{gray!20}{34.1} & 27.5 & 25.5 & 51.0 & 29.0 & 39.0 & 48.0 & 40.5 & 38.0 & 22.5 & 22.5 & 43.0 & 34.0 & 22.5 & 32.5 & 45.5 & \textbf{32.5} & 40.0 & 30.0 & 21.0 & 37.0 \\
        VideoChat~\cite{li2023videochat} & Vicuna-7B & \cellcolor{gray!20}{35.5} & 33.5 & 26.5 & 56.0 & 33.5 & 40.5 & 53.0 & 40.5 & 30.0 & 25.5 & 27.0 & 48.5 & 35.0 & 20.5 & 42.5 & 46.0 & 26.5 & \textbf{41.0} & 23.5 & 23.5 & 36.0 \\
        {VideoChat2~\cite{li2023mvbench}} & Vicuna-7B & \textcolor{gray!80}{{51.1}} & \textcolor{gray!80}{{66.0}} & \textcolor{gray!80}{{47.5}} & \textcolor{gray!80}{{83.5}} & \textcolor{gray!80}{{49.5}} & \textcolor{gray!80}{{60.0}} & \textcolor{gray!80}{{58.0}} & \textcolor{gray!80}{{71.5}} & \textcolor{gray!80}{{42.5}} & \textcolor{gray!80}{{23.0}} & \textcolor{gray!80}{{23.0}} & \textcolor{gray!80}{{88.5}} & \textcolor{gray!80}{{39.0}} & \textcolor{gray!80}{{42.0}} & \textcolor{gray!80}{{58.5}} & \textcolor{gray!80}{{44.0}} & \textcolor{gray!80}{{49.0}} & \textcolor{gray!80}{{36.5}} & \textcolor{gray!80}{{35.0}} & \textcolor{gray!80}{{40.5}} & \textcolor{gray!80}{{65.5}} \\

        \hline
 \multicolumn{22}{l}{\textit{\textbf{Text-Only Pre-Alignment Video-MLLM:}~TOPA zero-shot inference with 10 frames.}}   \\
        TOPA-ZeroShot&LLama2-7B&\cellcolor{gray!20}{39.8}&\textbf{42.0}&38.5&35.0&34.5&66.0&52.5&47.5&28.0&22.0&37.5&\textbf{81.0}&38.0&24.0&42.5&41.5&28.5&34.0&23.5&49.0&30.5\\

        TOPA-ZeroShot&LLama2-13B&\cellcolor{gray!20}\textbf{42.5}&38.0&\textbf{40.0}&42.5&\textbf{35.0}&\textbf{69.0}&52.5&\textbf{58.5}&29.5&22.5&\textbf{43.5}&80.5&38.0&25.5&\textbf{43.0}&{43.0}&29.5&37.5&\textbf{38.5}&\textbf{50.0}&32.5\\

        \hline
        \end{tabular}
    }
    \caption{
    {Evaluation results on MVBench. The results of other approaches are sourced from ~\cite{li2023mvbench}. We gray out the results of VideoChat2 since it utilizes extensive annotated downstream video data.}
    }
    \label{tab:mvpbench}
\end{table}

\begin{table}[h]
    \small
    \centering
    \setlength\tabcolsep{1.5pt}
    \resizebox{1.0\textwidth}{!}{
        \begin{tabular}{c|c|c|c|c|c|c|c|c|c}
        \hline
        \textbf{AS} & \textbf{AP} & \textbf{AA} & \textbf{FA} & \textbf{UA} & \textbf{OE} & \textbf{OI} & \textbf{OS} & \textbf{MD} & \textbf{AL}\\

        {Action} & {Action} & {Action} & {Fine-grained} & {Unexpected} & {Object} & {Object} & {Object} & {Moving} & {Action}  \\

        {Sequence} & {Prediction} & {Antonym} & {Action} &{Action} & {Existence} & {Interaction} & {Shuffle} & {Direction} & {Localization} \\

        STAR~\cite{star}&STAR~\cite{star}&PAXION~\cite{paxion}&MiT V1~\cite{mit}&FunQA~\cite{funqa}&CLEVRER~\cite{clevrer}&STAR~\cite{star}&Perception
Test~\cite{perception_test}&CLEVRER~\cite{clevrer}&Charades-STA~\cite{charades_sta}\\
    \hline
\textbf{ST} & \textbf{AC} & \textbf{MC} & \textbf{MA} & \textbf{SC} & \textbf{FP} & \textbf{CO} & \textbf{EN} & \textbf{ER} & \textbf{CI} \\

        {Scene} & {Action} & {Moving} & {Moving} & {State} & {Fine-grained} & {Character} & {Egocentric} & {Episodic} & {Counterfactual} \\

        {Transition} & {Counting} & {Counting} & {Attribute} & {Change} & {Pose} & {Order} & {Navigation} & {Reasoning} & {Inference} \\
        MovieNet~\cite{movienet}&Perception Test~\cite{perception_test}&CLEVRER~\cite{clevrer}&CLEVRER~\cite{clevrer}&Perception Test~\cite{perception_test}&NTU RGB+D~\cite{ntu_rgbd}&Perception Test~\cite{perception_test}&VLN-CE~\cite{scalevln}&TVQA~\cite{tvqa}&CLEVRER~\cite{clevrer}\\
        \hline
        \end{tabular}
    }

\end{table}

\subsection{Supervised Finetuning} \label{exp:finetuning}
In this section, we further finetune the pre-aligned TOPA models to study the benefits of TOPA for downstream supervised learning. 
During finetuning, TOPA directly takes the video feature as input without the cross-modal projection. 
More finetuning details for each dataset are provided in Appendix~\ref{app:training_details}.
Table~\ref{tab:fintune} shows the finetuning results on multi-choice video QA dataset. 
For comparison, we include baseline models without text-only pretraining. 
Our text-only pre-alignment consistently improves the performance across three benchmarks. Notably, TOPA-Llama2-7B achieves 67.1\% accuracy on TVQA, outperforming other approaches by a large margin. 
These results suggest that our text-only pre-alignment, even without training on real videos, has a similar effect to conventional video-language pre-training.
\begin{table}[ht]
\small
\vspace{-1em}
\caption{Finetuning results on NExT-QA, STAR and TVQA.}
\label{tab:fintune}
\centering
\setlength{\tabcolsep}{1.8mm}
\begin{tabular}{l ccccccccccc}
\tline
\multirow{2}{*}{\textbf{Model (\# Frames)}} & \multicolumn{4}{c}{\textbf{NExT-QA}} & \multicolumn{5}{c}{\textbf{STAR}} & \multirow{2}{*}{\textbf{TVQA}}  \\

\cline{2-5}
\cline{6-10}
& {Tem.} & {Cau.} & {Des.} & \textbf{Avg.} & {Int.} & {Seq.} & {Pre.} & {Fea.} & \textbf{Avg.} &\\
\hline

FrozenBiLM (10)~\cite{frozenbilm}  & -& -& - & - & - & - & - & - & - & 57.5 \\

InternVideo (8)~\cite{wang2022internvideo}  & 58.5  & 62.5 & 75.8 & 63.2 & 62.7 & 65.6 & 54.9 & 51.9 & 58.7 & 57.2 \\
BLIP-2$^{\text{voting}}$ (4)~\cite{sevila}  & 65.2 & 70.1 & 80.1 & 70.1 & 52.3 & 54.8 & 49.0& 51.2 & 51.8  & 54.5 \\ 
\sevila{} (32 $\rightarrow$ 4)~\cite{sevila}& {69.4} & {74.2}  & {81.3}  & {73.8} & {63.7} & {70.4} & {63.1} & {62.4} & {64.9}  & {61.6} \\ 
Llama-VQA-7B (10)~\cite{LLamavqa}~~ &69.2&72.7&75.8&72.0&66.2&67.9&57.2&52.7&65.4&-\\
\midrule
Baseline~(10)  &65.3&69.0&72.6&68.4&60.8&61.5&49.2&49.8&59.4&63.8\\
TOPA-Llama2-7B~(10)      &71.3&74.2&78.5&73.9&66.8&68.9&59.1 &55.5&66.4&67.1\\
                &\textcolor{blue}{+6.0}&\textcolor{blue}{+5.2}&\textcolor{blue}{+5.9}&\textcolor{blue}{+5.5}&\textcolor{blue}{+6.0}&\textcolor{blue}{+7.4}&\textcolor{blue}{+9.9}&\textcolor{blue}{+5.7}&\textcolor{blue}{+7.0}&\textcolor{blue}{+3.3}\\

\cdashline{1-12}
       Baseline~(10)&66.0&69.7&73.7&69.1&61.4&62.4&50.6&51.8&60.3&66.2\\
TOPA-Llama3-8B~(10) &70.1&74.5&74.6&73.1&66.3& 67.0 & 59.1 &56.5&65.4&68.1\\

&\textcolor{blue}{+4.1}&\textcolor{blue}{+4.8}&\textcolor{blue}{+0.9}&\textcolor{blue}{+4.0}&\textcolor{blue}{+4.9}&\textcolor{blue}{+4.6}&\textcolor{blue}{+8.5}&\textcolor{blue}{+4.7}&\textcolor{blue}{+5.1}&\textcolor{blue}{+1.9}\\

\cdashline{1-12}
Baseline~(10) &67.8&71.6&75.2&70.9&58.7&59.5&54.3&51.8&58.2&66.6\\
TOPA-Llama2-13B~(10)      &72.1&75.8&79.3&75.1&66.8&68.3&61.0&55.1&66.3&69.0\\
&\textcolor{blue}{+4.3}&\textcolor{blue}{+4.2}&\textcolor{blue}{+4.1}&\textcolor{blue}{+4.2}&\textcolor{blue}{+8.1}&\textcolor{blue}{+8.8}&\textcolor{blue}{+6.7}&\textcolor{blue}{+3.3}&\textcolor{blue}{+8.1}&\textcolor{blue}{+2.4}\\
\tline

\vspace{-1em}
\end{tabular}
\end{table}

\begin{figure*}[ht]
    \centering
    \includegraphics[width=1.0\textwidth]{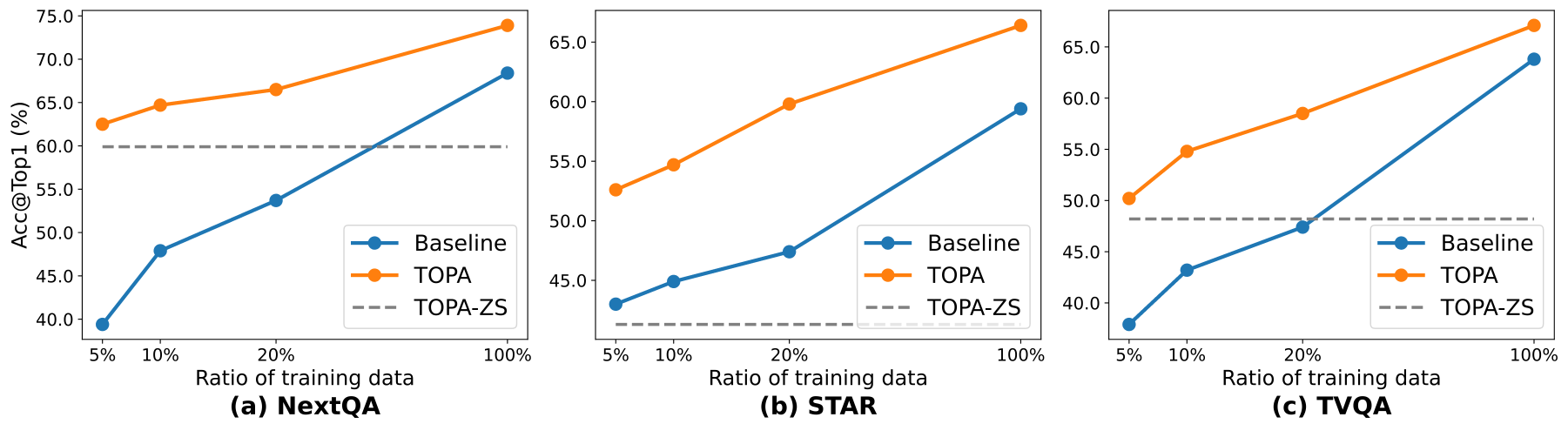}
    \vspace{-1em}
    \caption{Results of finetuning TOPA with various ratios of training data.}
    \label{fig:dataeff}
    \vspace{-0.1in}
\end{figure*}

\textbf{Data-efficient finetuning.}
Figure~\ref{fig:dataeff} shows the results of finetuning LLMs with various ratios of training data. TOPA trained with 10\% data achieves 64.7\% Top 1 accuracy on NeXT-QA benchmark, significantly outperforming the baseline that without text-only pre-alignment. Besides, when trained with less than 20\% data, the baseline model even performs worse than TOPA-zeroshot on NeXT-QA and TVQA, clearly demonstrating the effectiveness of TOPA in limited annotated data scenarios.

\subsection{Video Captioning} \label{exp:video captioning}
\textbf{Results on zero-shot video captioning.}
We further perform zero-shot video captioning on MSR-VTT~\cite{xu2016msrvtt} and VATEX~\cite{wang2019vatex}. As shown in Table~\ref{tab:captioning}, TOPA largely outperforms previous text-only approaches like Decap which is trained on captions sourced from CC3M~\cite{cc3m}. TOPA even outperforms the video-text pre-training approaches like VideoCoCa, which is pre-trained on millions of videos-text data, demonstrating that TOPA is an efficient and effective framework for video-LLM alignment.

\begin{table}[ht]
\small
\caption{Zero-shot video captioning results. We report CIDEr score for all benchmarks. \textit{VT} denotes {\pair{video clip}{text}} pairs, \textit{IT} denotes {\pair{image}{text}} pairs, and \textit{WP} denotes webpages consisting of interleaved image and text data.}
\label{tab:captioning}
\centering
\setlength{\tabcolsep}{1.5mm}
\begin{tabular}{lccc}
\tline
&\textbf{Training data}&\textbf{MSR-VTT}&\textbf{VATEX}\\
\hline
\multicolumn{2}{l}{\textit{(Web video-text Pre-training)}} \\
VideoCoCa-g~\cite{yan2022videococa}                     & 144M~\textit{{VT}}&27.1&22.8\\
Flamingo-3B ~\cite{alayrac2022flamingo}                 & 27M~\textit{VT}~\&~2.1B~\textit{IT}~\&~43M~\textit{WP}&-&40.1\\
Flamingo-9B ~\cite{alayrac2022flamingo}                 & 27M~\textit{VT}~\&~2.1B~\textit{IT}~\&~43M~\textit{WP}&-&39.5\\
VideoPrism-B~\cite{zhao2024videoprism} w/ PaLM-2-1B     & 618M~\textit{VT}&40.3&24.2\\
VideoPrism-B~\cite{zhao2024videoprism} w/ PaLM-2-8B~~~~ & 618M~\textit{VT}&38.5&31.7\\
\hline
\multicolumn{2}{l}{\textit{(Text-only Pre-training)}} \\
DeCap~\cite{li2023decap} &3M~\textit{Captions}&18.6&18.7\\
TOPA-Llama2-7B& 721K \textit{TextVid}&32.9&31.0\\
TOPA-Llama2-13B &721K \textit{TextVid}&33.4&32.0\\
\tline
\end{tabular}
\end{table}

\subsection{Ablations} \label{exp:ablations}
\textbf{LLM prior in video-language understanding.}
To investigate the impact of LLM prior in multi-choice video QA, we conduct experiments on EgoSchema with the blind setting, where only the questions and choices are provided to the LLM. Table~\ref{tab:blindVQA} shows the results.
Bard and GPT-4-Turbo achieve 33.2\% and 30.8\% accuracy, respectively. Gemini-Pro-1.0 reaches 38.2\% accuracy.
These blind results of advanced LLMs suggest that in some video QA cases, LLMs can accurately choose the correct answer solely based on the question and choices, without visual input.
However, the blind performance of Llama2-7B and Llama2-13B is inferior, potentially due to their smaller model size. 
After training on the TextVid dataset, TOPA-Llama2-13B achieves a blind accuracy of 37.5\% (or +11.7\%), closely approaching that of Gemini-Pro-1.0 model. These results suggest that text-only pre-alignment can effectively prepare LLMs for downstream video-language tasks by leveraging specialized text-only tasks, even in complex scenarios where the original LLMs are limited.

\begin{table*}[ht]
\vspace{-0.2cm}
\small
\setlength{\tabcolsep}{0.9mm}
\begin{floatrow}
\capbtabbox[7cm]{
    \vspace{-1em}
    \begin{tabular}{l cc}
    \tline
    &{Visual Input}& {~~ES Full~~}\\
    \hline
    Random Selection&\multirow{4}{*}{{\XSolidBrush}}&20.0 \\
    GPT-4-Turbo~$^\dagger$& &30.8 \\
    Bard~$^\dagger$&&33.2 \\
    Gemini-Pro-1.0  && 38.2 \\
    \cdashline{1-3}
    Llama2-7B~&\multirow{2}{*}{{\XSolidBrush}}&20.1 \\
    Llama2-13B~&&25.8 \\
    \cdashline{1-3}
    TOPA-Llama2-7B~&\multirow{2}{*}{{\XSolidBrush}}&29.3 \\
    TOPA-Llama2-13B~&&37.5 \\
    \hline
    TOPA-Llama2-7B  &\multirow{2}{*}{\ding{52}}&41.2 \\
    TOPA-Llama2-13B~~ &&51.0 \\
    \tline
    \end{tabular}
    }
{
 \caption{Blind results on EgoSchema. $\dagger$ denotes results sourced from \cite{MC:MemoryConsolidation}.}
 \label{tab:blindVQA}
}
\small
\setlength{\tabcolsep}{1mm}
\capbtabbox{
    \vspace{-1em}
    \begin{tabular}{lccc}
    \tline
    TOPA&\#Frame&NextQA&ES Full \\
    \hline
    \multirow{3}{*}{Llama2-7B}&1&56.1&39.4\\
    &5&58.9~(+2.8)&41.0~(+1.6) \\
    &10&59.9~(+3.8)&41.2~(+1.8) \\
    \hline
    \multirow{3}{*}{Llama2-13B}&1&57.3&47.6\\
    &5&60.8~(+3.5)&50.5~(+2.9)\\
    &10&62.1~(+4.8)&51.0~(+3.4) \\
    \tline
    \end{tabular}
    }
{
 \caption{Ablation on video frames.}
 \label{tab:singleframe}
 \small
}

\end{floatrow}
\end{table*}

\textbf{The impact of video frames.} 
To better investigate TOPA's capability in understanding temporal dynamics of real videos, we conduct experiments with different number of frames.  Table~\ref{tab:singleframe} shows the results.
Multiple frames input consistently enhances performance on NeXT-QA and EgoSchema for both TOPA-Llama2-7B and TOPA-Llama2-13B. 
This indicates that the text-only pre-alignment effectively enables the LLM to handle multiple video frames, despite not being trained on real videos.



\section{Conclusions}
In this paper, we introduce TOPA, a text-only pre-alignment framework designed for aligning LLMs with video modality without requiring training on real videos. TOPA has demonstrated remarkable performance on the recent, challenging long-form video understanding benchmark, \ie~EgoSchema, showcasing that a text-only approach is effective in capturing the dynamics of long-form videos. Our approach, which includes data generation and text-only pre-alignment, has potential applications across various vision-language tasks where obtaining paired vision-language data is difficult.

\newpage

\section*{Acknowledgements}
This work was supported by National Key R\&D Program of China (No. 2023YFC3305600), the National Natural Science Foundation of China (U2336212), the Fundamental Research Funds for the Zhejiang Provincial Universities (226-2024-00208), Lu's Graduate Education International Exchange Foundation and the National Research Foundation, Singapore under its Strategic Capability Research Centres Funding Initiative. Any opinions, findings and conclusions or recommendations expressed in this material are those of the author(s) and do not reflect the views of National Research Foundation, Singapore. The computational work was partially performed on resources of the National Supercomputing Centre, Singapore (https://www.nscc.sg).

\bibliography{neurips}

\newpage
\section*{Appendix}
In Appendix~\ref{app:additional experiments}, we provide additional experiments and analysis. 
\begin{itemize}

    \item In Appendix~\ref{app:MCQA}, we further discuss the Multi-choice QA task and study the impact of the multi-choice Tideo QA pre-training. 
    
    \item In Appendix~\ref{app:gap}, we study the impact of cross-modal projection~(Eq.~\ref{eq:project}). 

    \item In Appendix~\ref{app:Qualitative Results}, we provide extensive qualitative results to illustrate TOPA's advantages and limitations across various video understanding tasks.

\end{itemize}

Appendix~\ref{sec:limitation}: The limitations of TOPA.

Appendix~\ref{sec:boarder impact}: The broader impact of TOPA.

Appendix~\ref{app:dataset}: The details of proposed TextVid dataset.

Appendix~\ref{app:benchmarks}, The details of benchmarks.

Appendix~\ref{app:training_details}, The training details of TOPA.

Appendix~\ref{app:prompts}: The prompts used in this paper.

Appendix~\ref{sec:license}: The licenses of datasets, codes and models used in this paper.

Appendix~\ref{sec:examples}: Examples from TextVid.

\newpage

\appendix
\section{Additional Experiments} \label{app:additional experiments}

\subsection{Further Discussion on Multi-Choice Video QA Task} \label{app:MCQA}
\begin{table}[ht]
\small
\vspace{-0.3cm}
\caption{Multi-choice video QA on EgoSchema subset and full set. ``Gap'' refers to the difference in performance between the subset and the full set
}
\label{tab:ego_app}
\centering
\setlength{\tabcolsep}{2mm}
\begin{tabular}{l lcccc}
\tline
\textbf{Method}&\textbf{Eval Mode}&\textbf{ES Subset}&\textbf{ES Full}&\textbf{Gap}\\
\hline 
LongViViT~\cite{longvivit}&Similarity&56.8&33.3&-23.5\\
MC-ViT-L~\cite{MC:MemoryConsolidation}&Similarity&62.6&44.0&-18.6\\
MVU~\cite{onepass}&LLM logits&60.3&37.6&-22.7\\
LangRepo-Mixtral-8×7B-(12B active)~\cite{langrepo}&LLM logits&66.2&41.2&-25.0\\
VideoAgent~(GPT-4)~\cite{agent_video}&LLM Selection&60.2&54.1&-6.1\\
\midrule
TOPA-LLama2-13B & LLM Logits &67.5&41.6&-25.9\\
TOPA-LLama2-13B & LLM Selection &51.2&51.0&-0.2\\
\midrule
TOPA-LLama2-7B & LLM Logits &64.5&41.7&-22.8\\
TOPA-LLama2-7B & LLM Selection &40.4&41.2&+0.8\\
\midrule
TOPA-LLama2-7B (w/o multi-choice training) & LLM Logits &65.1&40.5&-24.6\\
TOPA-LLama2-7B (w/o multi-choice training) & LLM Selection &24.3&24.7&+0.4\\

\tline
\end{tabular}
\end{table}
A significant advantage of the text-only framework is that we can utilize the LLM to automatically generate diverse language-based supervisions as needed, such as the multi-choice QA pairs. To explore the impact of the multi-choice QA training tasks, we conduct an ablation study as shown in Table~\ref{tab:ego_app}. 
We would like to first introduce the different evaluation modes for multi-choice video QA tasks: (1) \textbf{LLM Selection}: Asking the LLM to predict the correct answer given the video-question-choices.
(2) \textbf{LLM Logits}: Given the video and question as LLM context, we calculate the logits for each choice by averaging the logits of all words within the choice. The choice with higher logit tends to match the video-question context better and is thus selected as the predicted answer.
(3) \textbf{Similarity Comparison}~\cite{longvivit,MC:MemoryConsolidation}: Mapping the multiple question-choice pairs and video to a common feature space and calculating the similarity between the video and each question-choice.

\textbf{The performance gap between the EgoSchema subset and full set.}
Previous work~\cite{MC:MemoryConsolidation,longvivit} highlights a huge performance gap between the subset and the full set of EgoSchema as shown in Table~\ref{tab:ego_app}. 
While concurrent work~\cite{langrepo,onepass} introduces log-likelihood based approaches for LLM inference, which significantly improve the performance on EgoSchema subset, the issue of the performance gap still persists.
In this paper, we observe that such a performance gap phenomenon also occurs in approaches based on LLM logits. However, but it diminishes or even disappears in methods employing LLM selection.
We find that this phenomenon may be attributed to differences in the linguistic structures of the choices, as shown below. The choices in the subset often differ in several key works like ``create'', ``repair'' and ``clean''. 
The similarity or logit can effectively identify this keyword-level difference to select a more appropriate choice.
Conversely, the choices in the full set display more substantial linguistic differences. These variations introduce significant language biases, \ie~some sentences naturally receive higher logits in LLM, complicating the reliance on similarity or logit for choice selection. 
In contrast, LLM selection methods take all the choices within the context, allowing the LLM to leverage its robust contextual understanding to select the correct choice.

\colorbox{gray!10}{%
\begin{minipage}{0.96\textwidth}
\small{{\textcolor{navy}{
Question-Choices examples from subset:
}}}

\small{\textit{\textcolor{navy}{
Q: Can you summarize the primary objective and the steps the person took throughout the video to achieve it? ensure your answer captures the essence of the video without listing all actions.
}}}

\small{\textit{\textcolor{navy}{
A: The main aim of the person's primary objective was to \textbf{create and build} a new, sturdy wooden bench.
}}}

\small{\textit{\textcolor{navy}{
B: The primary objective for the person was to thoroughly \textbf{repair and restore} the wooden bench.
}}}

\small{\textit{\textcolor{navy}{
C: The person's primary objective was to thoroughly \textbf{clean and sanitize } the wooden bench's surface.
}}}
 \\

\small{{\textcolor{maroon}{
Question-Choices examples from full set:
}}}

\small{\textit{\textcolor{maroon}{
Q: Considering the entire video, what would you identify as the most crucial moments in the person's shopping experience and why?
}}}

\small{\textit{\textcolor{maroon}{
A: Following a strict shopping list as a guideline and rejecting unfit produce.
}}}

\small{\textit{\textcolor{maroon}{
B: Conducting taste tests and checking for the freshness of each vegetable.
}}}

\small{\textit{\textcolor{maroon}{
C: Using math algorithm for optimal vegetable selection.
}}}
\end{minipage}
}

\textbf{LLM for multi-choice QA.}
In Table~\ref{tab:ego_app}, we observe a notable phenomenon where the TOPA models achieve impressive results on the subset with the logits evaluation mode. TOPA-LLama2-13B achieves 67.5\% top1 accuracy, surpassing GPT-4-based video agents. However, when evaluated with the multi-choice selection mode, the performance of the subset declines to 51.2\%, but the performance of the full set increases from 41.6\% to 51.0\%. 
These results suggest that 
while the LLM is capable of selecting the answer from multiple choices, it is less sensitive to the keywords within those choices. In contrast, the logit-based approach is sensitive to the keywords but has difficulty with complex sentence understanding.

\textbf{The impact of the Multi-Choice Tideo QA pre-training.}
In Table~\ref{tab:ego_app}, we report the results of TOPA without the multi-choice Tideo QA task, \ie~trained with {\Tideo} summarization and {\Tideo} QA tasks. In this case, we find that TOPA-LLama2-7B maintains similar performance when evaluated with the logit mode. However, there is a significant performance drop when evaluated with the multi-choice selection mode. 
This result suggests that while the LLM is adapted to process video inputs, its capability is somewhat constrained and can not extend to more complex video-language tasks beyond the pre-training tasks. This finding highlights the advantage of our text-only data generation and text-only pre-alignment framework, which enable us to develop a variety of pre-alignment tasks to better equip the LLM for general video-language tasks such as dense captioning, multi-choice video QA, and video chat.

\subsection{The CLIP Modality Gap} \label{app:gap}
TOPA is pretrained with CLIP text features while inferenced with CLIP image features. We employ a modality projection approach, \ie~Eq.~\ref{eq:project}, to bridge this CLIP modality gap during zero-shot inference. Table~\ref{tab:project} shows the impact of Eq.~\ref{eq:project}. TOPA shows inferior results when directly taking the visual feature as input due to the modality gap problem.
The projection approach effectively alleviates such a modality gap problem without additional training.
\begin{table}[h]
\small
\caption{Ablation on the modality projection~(Eq.~\ref{eq:project}). Results on EgoSchema full set.
}
\label{tab:project}
\centering
\setlength{\tabcolsep}{1mm}
\begin{tabular}{lcc}
\tline
\textbf{Model}&\textbf{without Eq.~\ref{eq:project}}&\textbf{with Eq.~\ref{eq:project}}\\
\hline
TOPA-LLama2-7B&30.6&41.2\\
TOPA-LLama2-13B&38.3&51.0\\
\tline
\end{tabular}
\vspace{-0.5cm}
\end{table}

\clearpage
\subsection{Qualitative Results and Analysis} \label{app:Qualitative Results}
\begin{figure*}[ht]
    \centering
    \includegraphics[width=1.0\textwidth]{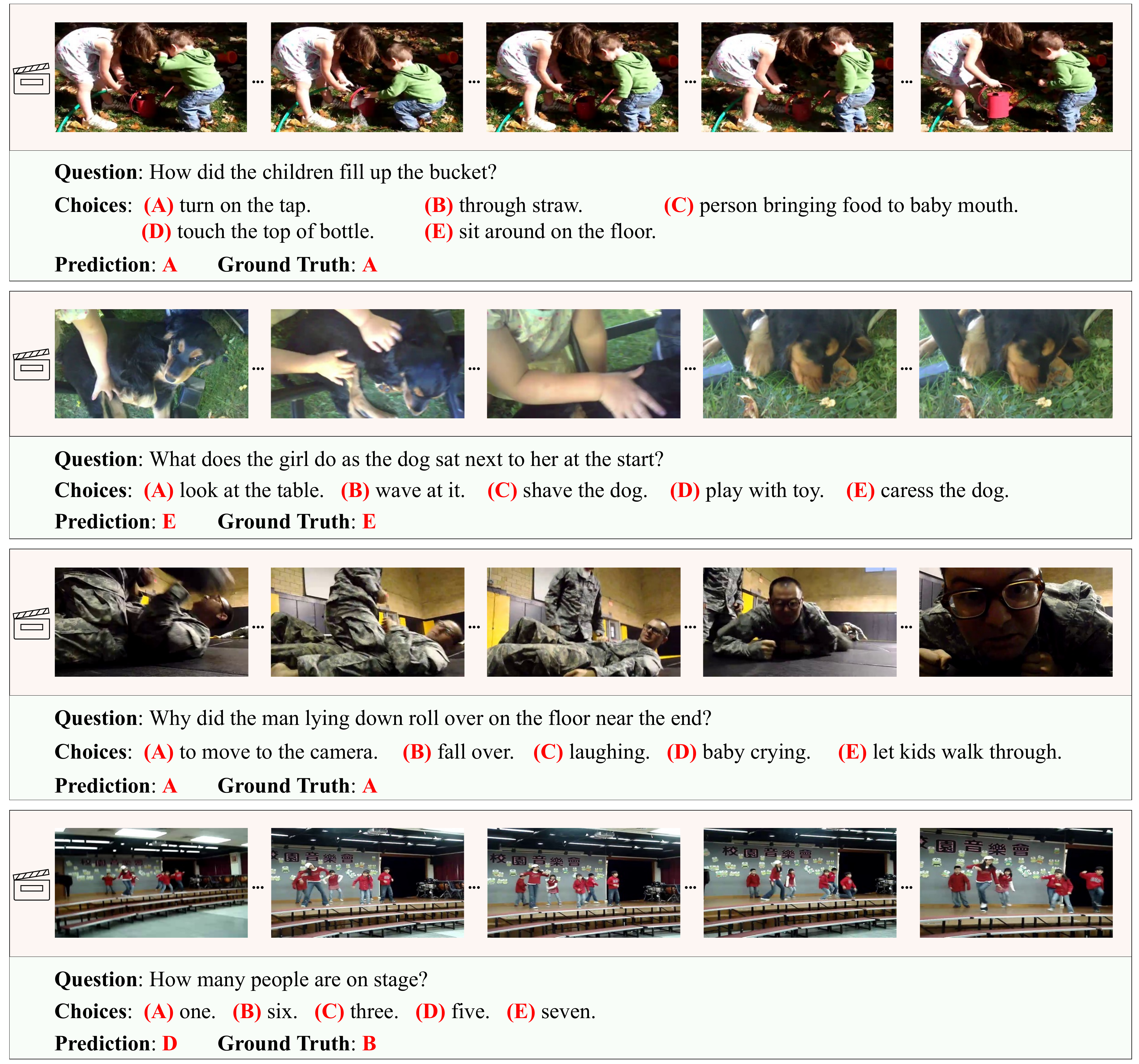}
    \vspace{-1em}
    \caption{Qualitative results on NeXT-QA. TOPA effectively performs complex video understanding tasks. Additionally, a failure case is also shown in the figure, \ie~in the last sample, TOPA failed to accurately count the number of people. }
    \label{fig:qualitatvie_nextqa}
    \vspace{-0.1in}
\end{figure*}

\begin{figure*}[ht]
    \centering
    \includegraphics[width=1.0\textwidth]{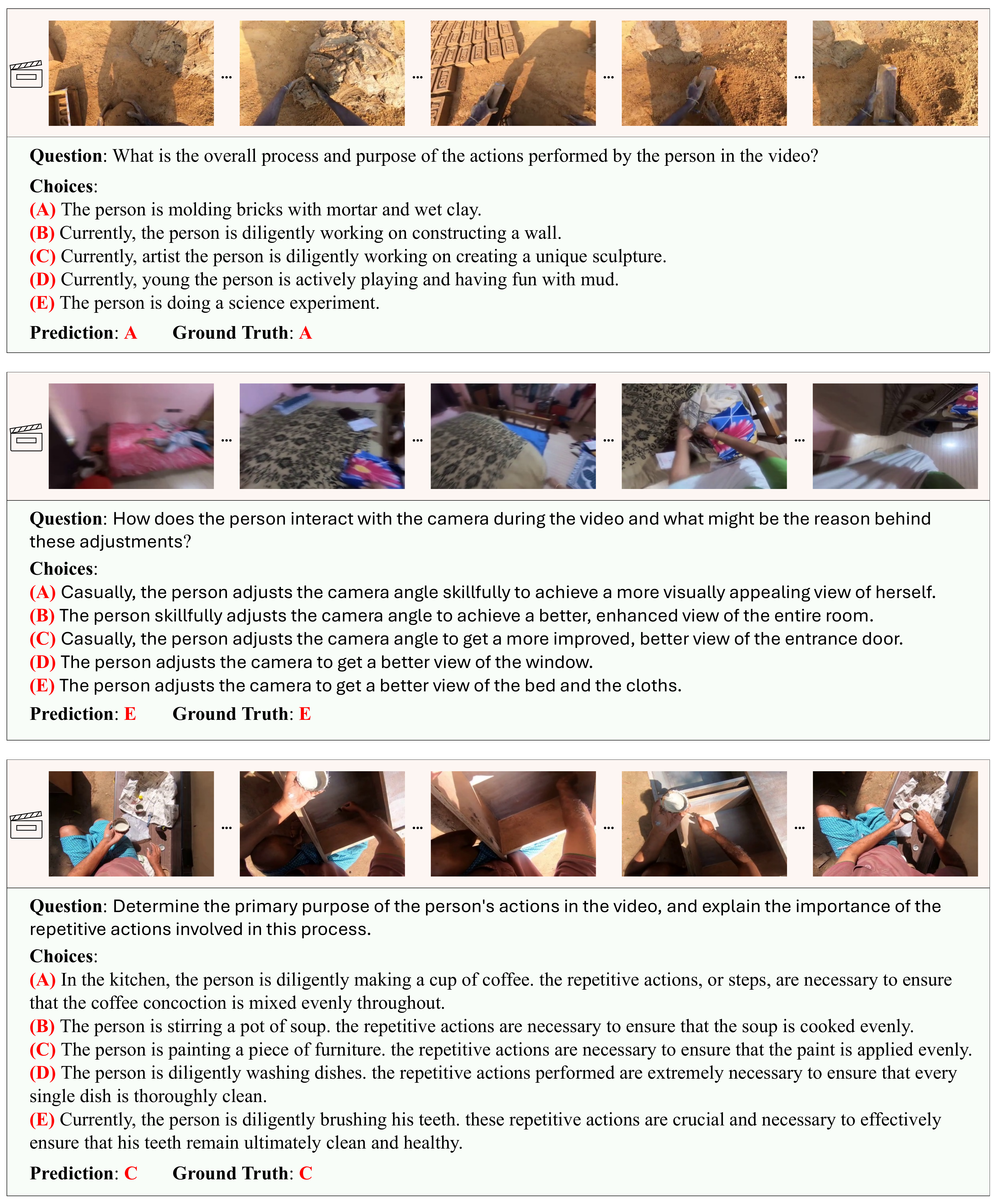}
    \caption{EgoSchema presents unique challenges compared to previous video benchmarks. The questions in EgoSchema are complex and demand advanced video capabilities, encompassing both recognition and reasoning skills.}
    \label{fig:qualitatvie_egos}
\end{figure*}

\begin{figure*}[ht]
    \centering
    \includegraphics[width=1.0\textwidth]{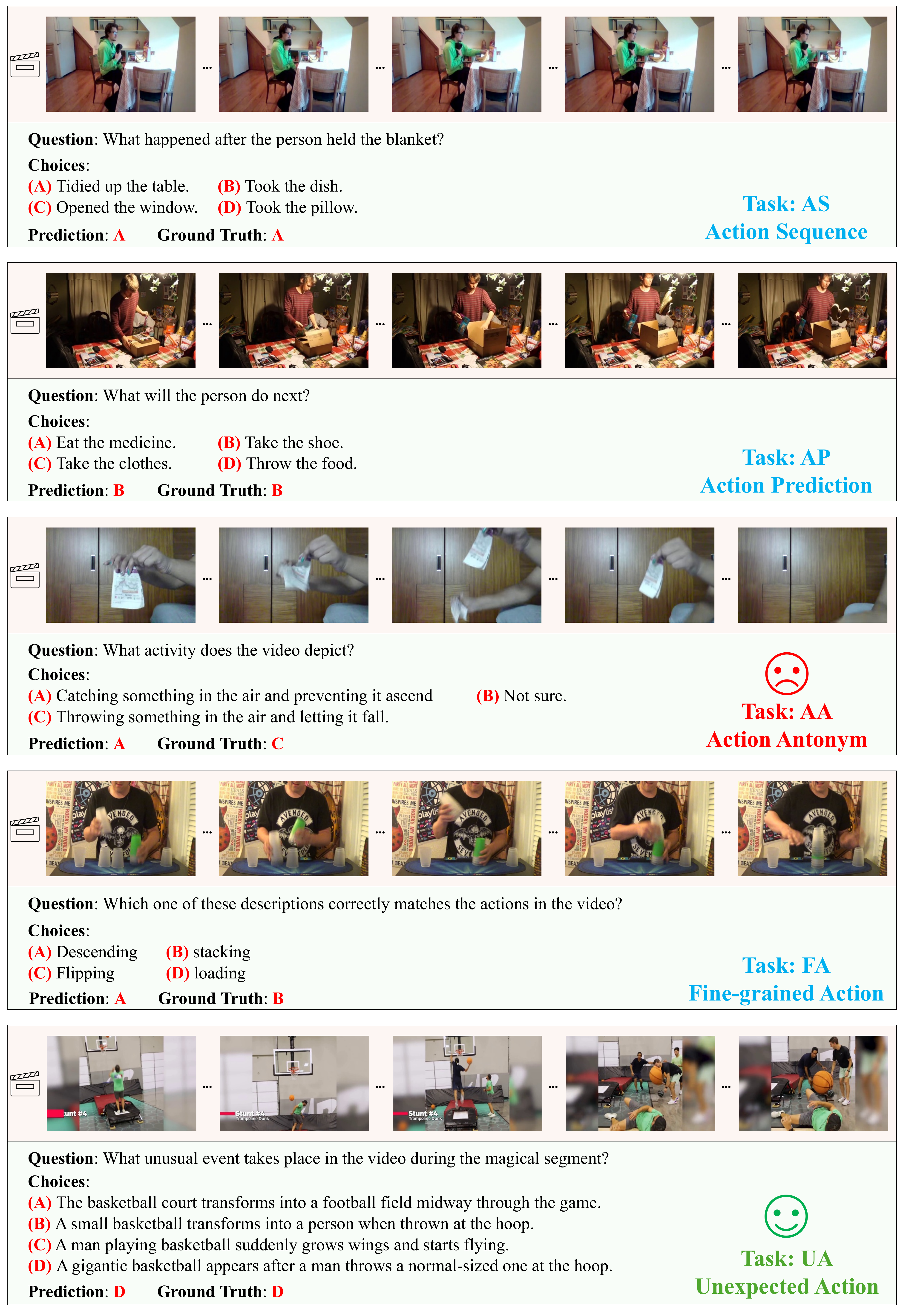}
    \vspace{-1em}
    \caption{Qualitative results on MVBench (Task 1-5). The tasks where TOPA performs well, average, or poorly are marked in green, blue, and red colors respectively.}
    \label{fig:mvbench1}
    \vspace{-0.1in}
\end{figure*}

\begin{figure*}[ht]
    \centering
    \includegraphics[width=1.0\textwidth]{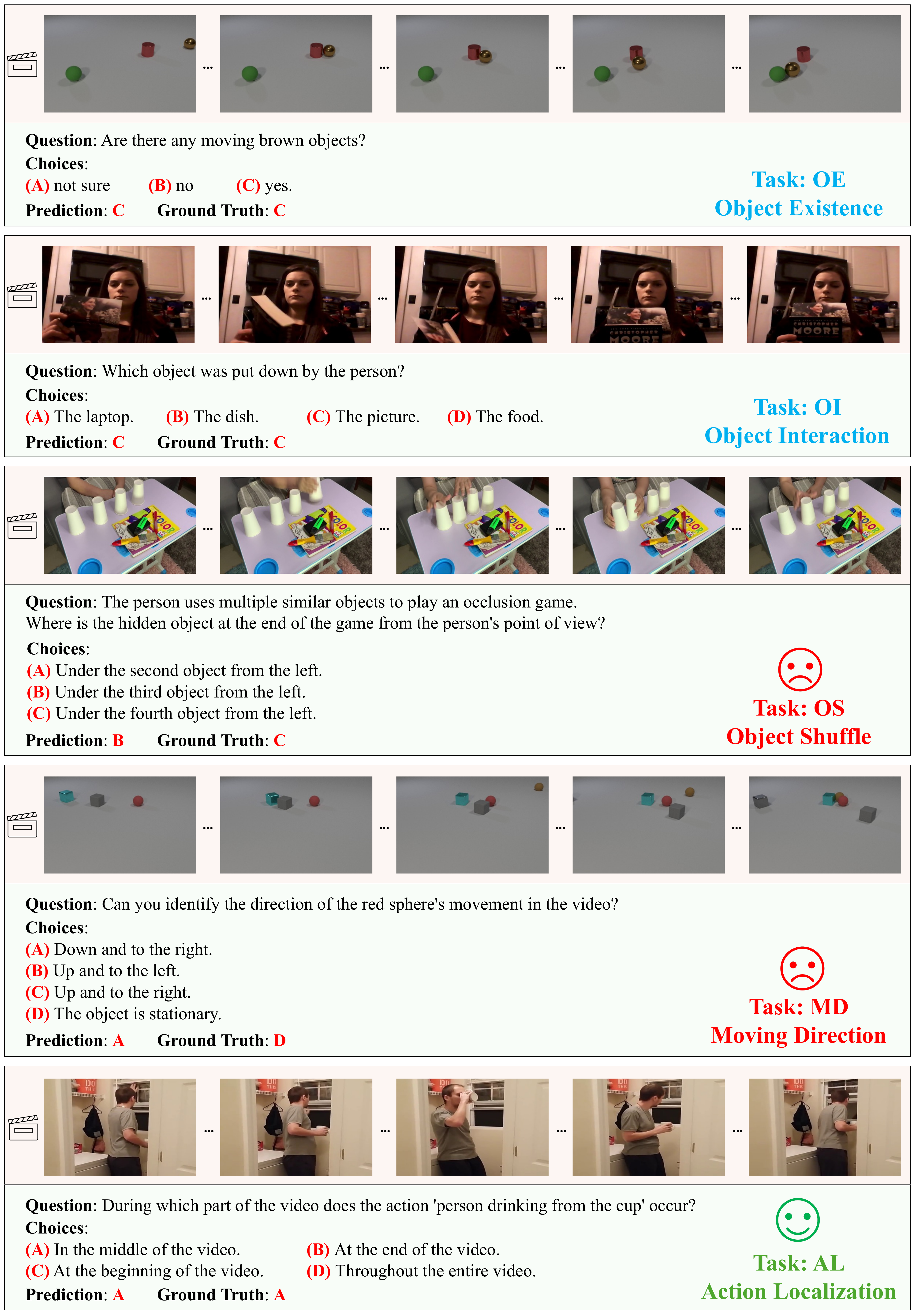}
    \vspace{-1em}
    \caption{Qualitative results on MVBench (Task 6-10). }
    \label{fig:mvbench2}
    \vspace{-0.1in}
\end{figure*}

\begin{figure*}[ht]
    \centering
    \includegraphics[width=1.0\textwidth]{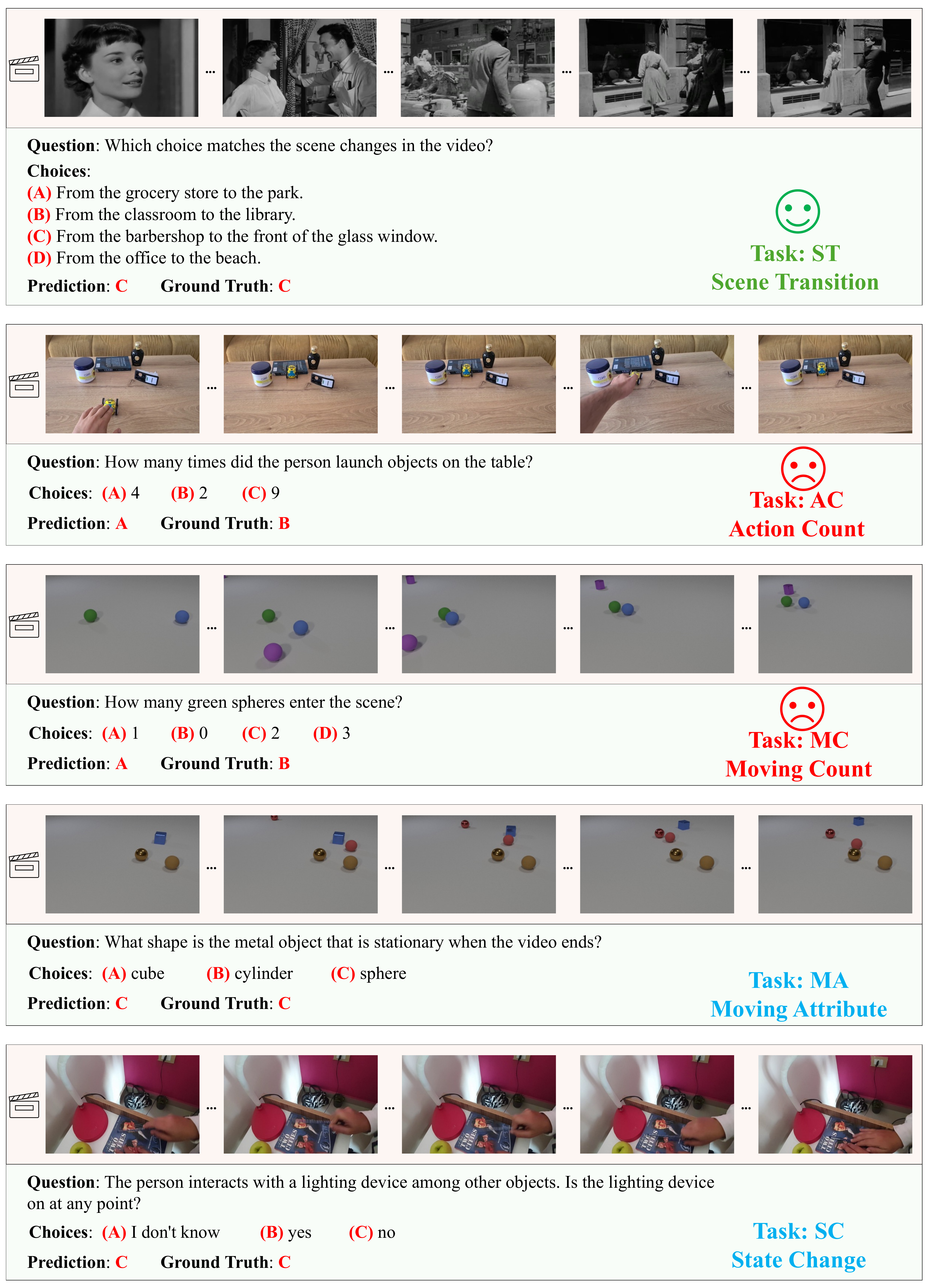}
    \vspace{-1em}
    \caption{Qualitative results on MVBench (Task 11-15). }
    \label{fig:mvbench3}
    \vspace{-0.1in}
\end{figure*}

\begin{figure*}[ht]
    \centering
    \includegraphics[width=1.0\textwidth]{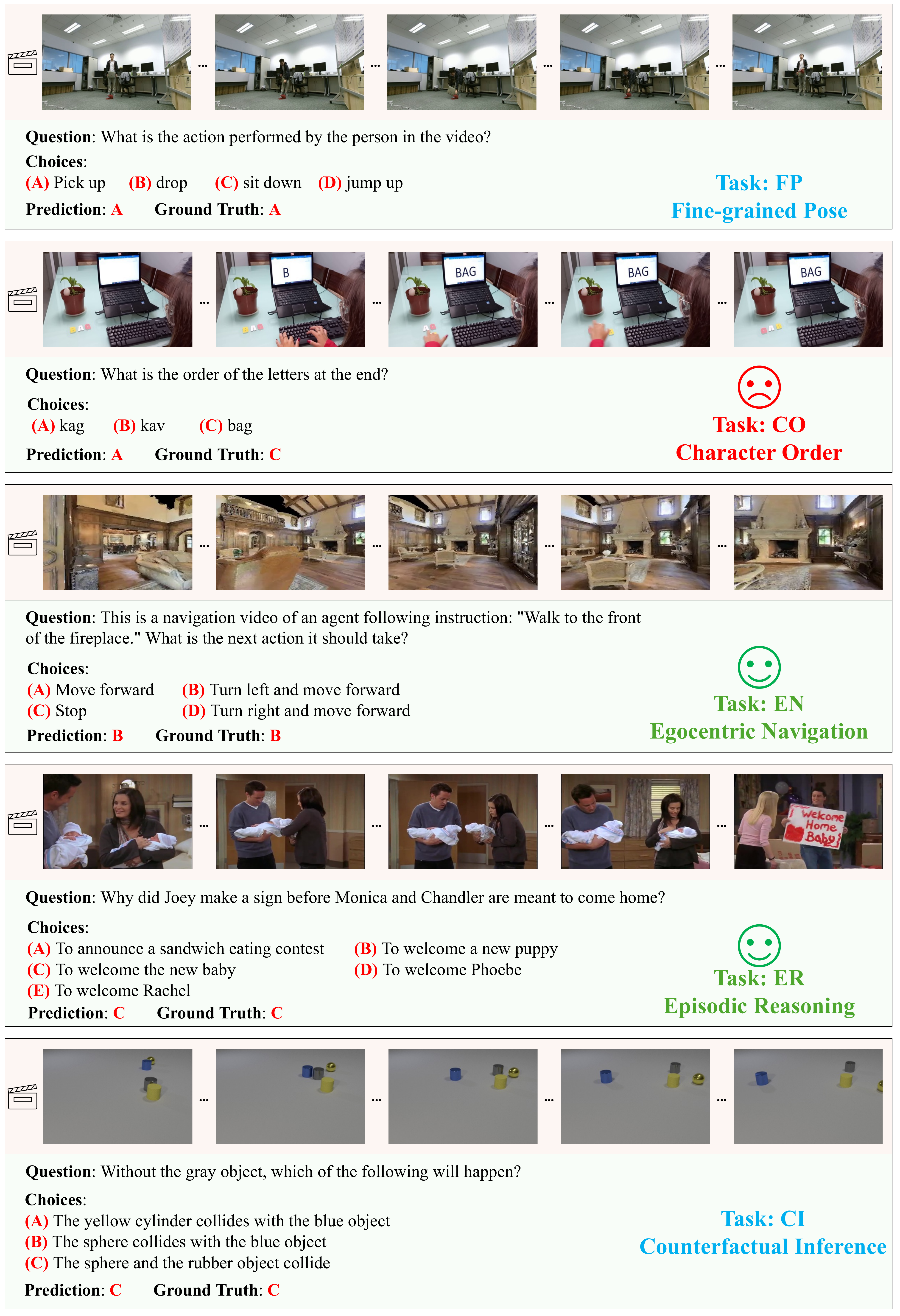}
    \vspace{-1em}
    \caption{Qualitative results on MVBench (Task 15-20). }
    \label{fig:mvbench4}
    \vspace{-0.1in}
\end{figure*}

We present qualitative results to illustrate the capabilities and limitations of TOPA across various video understanding tasks.  Figure~\ref{fig:qualitatvie_nextqa} shows qualitative results on the NExT-QA validation set. Figure~\ref{fig:qualitatvie_egos} shows qualitative results on the EgoSchema subset. 
Figure~\ref{fig:mvbench1}~-~\ref{fig:mvbench4} shows qualitative results on 20 video understanding tasks from MVBench.

\clearpage
\section{Limitations.} \label{sec:limitation}
\textbf{Modality gap in CLIP.} Despite the fact that TOPA achieves impressive results, a significant limitation in TOPA is the gap between the CLIP text feature and CLIP image feature. On the one hand, we use the CLIP text feature for pre-alignment, while inference is with the CLIP visual feature. The modality gap makes the performance degrade, despite employing a modality projection mechanism to mitigate it. On the other hand, the CLIP text features cannot fully capture the fine-grained visual details present in actual images, such as object locations and relationships. This limitation causes TOPA to struggle in scenarios where questions involve detailed visual information, as shown in Appendix~\ref{app:Qualitative Results}.

\textbf{Struggles in fine-grained visual understanding.} In TOPA, we propose textual videos to mimic real videos. However, this approach primarily focuses on keyframes understanding, which is insufficient for scenarios requiring the model to process hundreds of frames at high fps, such as action counting tasks. 
Besides, for the fine-grained action understanding scenarios, TOPA is unable to capture the fine-grained visual information. For example, in a scene where a person climbs a ladder, it is difficult for TOPA to identify whether the person is going up or down due to the limited capability to capture detailed visual dynamics.
Further enhancing TOPA with video instruction tuning might address these limitations which we leave for future work.



\section{Broader Impact} \label{sec:boarder impact}
\textbf{Academic Impact.} TOPA's methodology, which frees the need for costly video-text data collection and large-scale pre-training, lowers the barriers to entry for research and development in video-language understanding technologies. The text-only learning framework of TOPA may inspire researchers with limited resources to engage in cutting-edge multi-modal research, providing a more diverse range of perspectives and contributions to this field.

\textbf{Social Impact.} 
The ultimate objective of TOPA is to develop a general video-language understanding model. Its primary application enables users to extract information from long-form videos without the need for detailed viewing. Moreover, these capabilities for interpreting and managing video content could significantly enhance content moderation systems. Platforms hosting user-generated content could employ sophisticated video-language models to efficiently detect and mitigate the effects of inappropriate or harmful video content. 

\section{The details of proposed TextVid dataset} \label{app:dataset}

We utilize Gemini Pro 1.0 API for our data generation process. We prompt the LLM to create textual videos along with associated annotations. 
To ensure a diverse dataset that covers a wide of domains, we add condition prompts including different themes, video captions, video events, and the names of main objects.
Specifically, we leverage video titles from Howto100M~\cite{miech2019howto100m}, video captions from WebVid2M~\cite{webvid}, video events from Ego4D~\cite{ego4d}, and object from Wordnet~\cite{miller1995wordnet} as conditions to generate diverse textual videos. For Ego4D condition, we ask the LLM to mimic an ego-centric video to further improve the diversity of the dataset. Table~\ref{tab:textvid} compares vocabulary sizes.
Figure~\ref{fig:tideo_feature} shows that Tideos generated under different conditions have different distributions.
For each data generation, we prompt the LLM with the task prompt and one of the condition prompts as shown in Figure~\ref{fig:prompts}. 
The statistics TextVid are shown in Table~\ref{tab:textvid}. 
Additionally, we provide Wordcloud of TextVid in Figure~\ref{fig:wordcloud}. The examples of TextVid are shown in Appendix~\ref{sec:examples}.

\begin{table}[ht]
\small
\vspace{-0.1in}
\caption{Statistics of TextVid. 
}
\label{tab:textvid}
\centering
\setlength{\tabcolsep}{1mm}
\begin{tabular}{l|c}
\tline
&\textbf{TextVid}  \\
\hline
Generated by&Gemini Pro 1.0 \\
\hline
{\# textual videos}&721K       \\
\# each condition &Video Title-213k;~Video Caption-183K;~Video Scenarios-205K; Object-120K\\
\hline
\# all QA pairs &  3.5M         \\
\# each question types&What-2.5M;~Why-410K;~How-287K;Others-254K\\
\hline
\# all frames& 4.4M          \\
Avg. frame& 6.13 \\

\tline

\end{tabular}
\end{table}

\begin{table}[H]
\label{tideo_vocab}
\centering
\begin{tabular}{lccccc}
\toprule
& \textbf{Howto100m} & \textbf{Ego4D}&\textbf{WebVid}&\textbf{WordNet} \\
\midrule
\textbf{Vocab Size}&17492&7320&15095&26486\\
\bottomrule
\end{tabular}
\caption{Vocabulary size of Tideos generated under different prompts. We randomly sampled 20,000 global captions from each type of Tideos for comparison.}
\end{table}

\begin{figure*}[ht]
    \centering
    \includegraphics[width=0.95\textwidth]{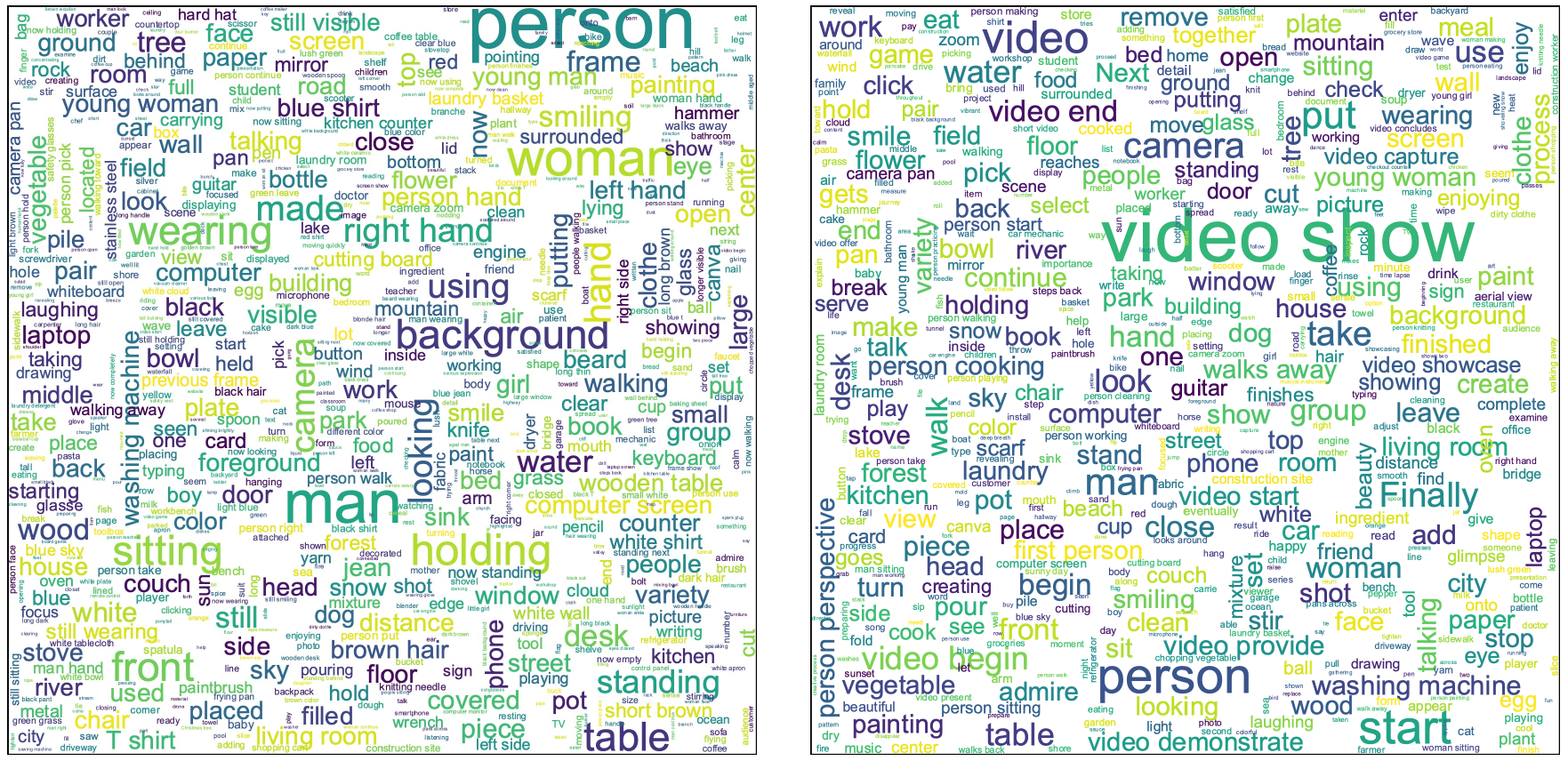}
    \vspace{-1em}
    \caption{Wordcloud of TextVid. The frame caption~(left) and the dense video caption~(right).}
    \label{fig:wordcloud}
\end{figure*}

\begin{figure}[h]
    \centering
    \includegraphics[width=0.8\linewidth]{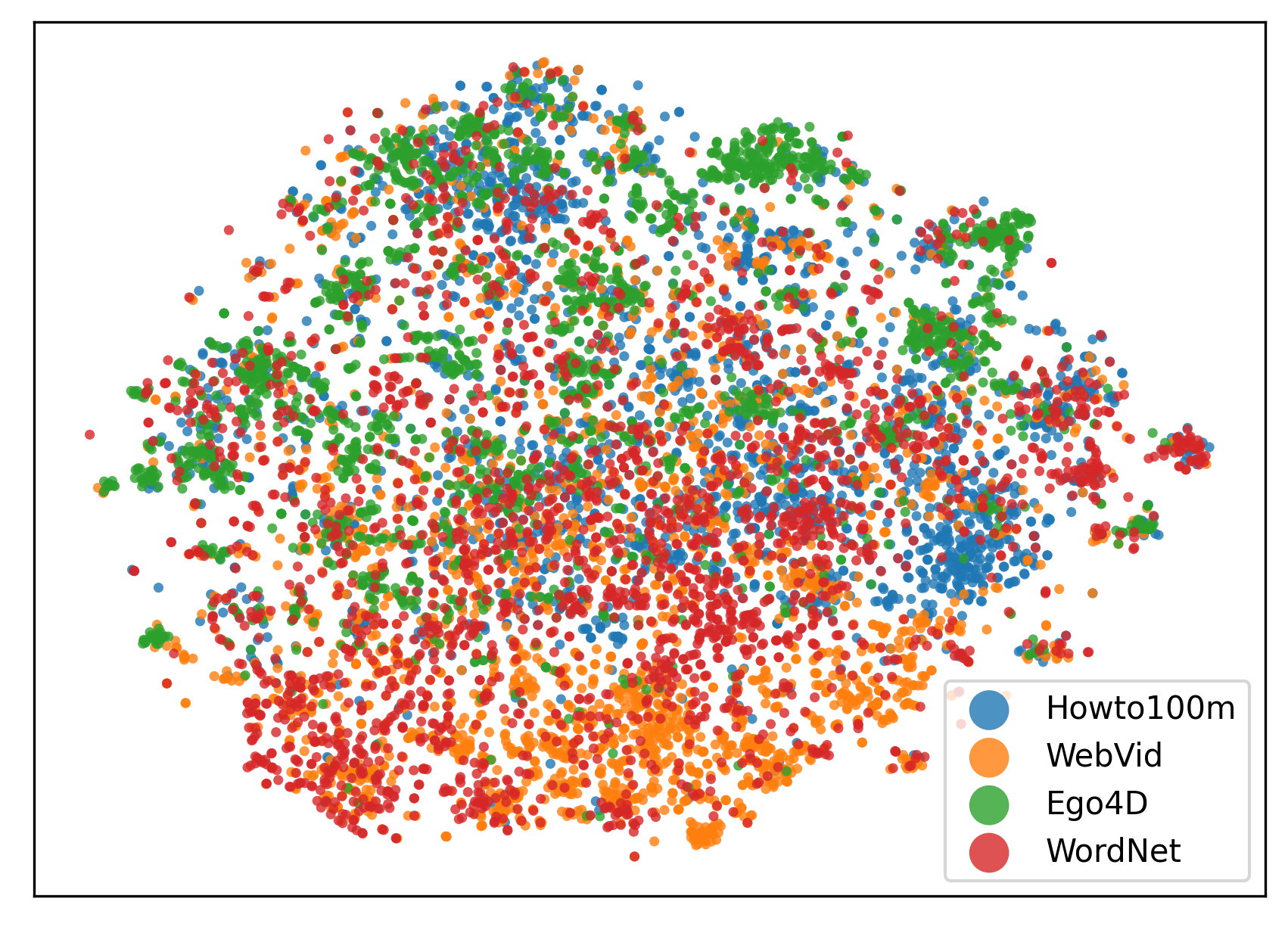}
    \caption{Visualization of Tideo features generated from different type of prompts.}
    \label{fig:tideo_feature}
\end{figure}

\section{Experimental Setup}
\subsection{Benchmarks} \label{app:benchmarks}
\textbf{EgoSchema}~\cite{EgoSchema} is a challenging long-form video understanding benchmark with 5000 multi-choice questions. The videos in EgoSchema are sourced from Ego-4D~\cite{ego4d}, with an average length of 3 minutes, distinct from previous benchmarks that focused on shorter, seconds-long videos. The questions in EgoSchema are manually curated to demand long temporal reasoning. We report results on EgoSchema full set.

\textbf{NExT-QA}~\cite{nextqa} is a multi-choice video QA benchmark for causal and temporal reasoning, including 5,440 natural videos. The average length of video is 44 seconds. We report results on NExT-QA validation set, which contains 570 videos and 5,000 multiple-choice questions.

\textbf{STAR}~\cite{star} is a benchmark for situated reasoning. It contains 22K video clips with an average length of 44 seconds. There are 4 different question types in STAR: Interaction (Int.), Sequence (Seq.), Prediction (Pre.), and Feasibility (Fea.). We report results on STAR validation set.

\textbf{TVQA}~\cite{tvqa} is a benchmark containing 21k video clips with an average length of 76 seconds.
We report results on TVQA validation set without subtitles.

\textbf{MVbench}~\cite{li2023mvbench} is a reorganized benchmark containing 20 video understanding tasks. These tasks are sourced from STAR~\cite{star}, PAXION~\cite{paxion}, MiT Vi~\cite{mit}, FunQA~\cite{funqa}, Perception Test~\cite{perception_test}, Charades-STA~\cite{charades_sta}, MovieNet~\cite{movienet}, NTU RGB+D~\cite{ntu_rgbd}, VLN-CE~\cite{scalevln} and TVQA~\cite{tvqa}.

\subsection{The details of training and evaluation.} \label{app:training_details}

We leverage Llama2-7B, Llama2-13B~\cite{llama2} and Llama3-8B as the LLM backbone. Additionally, we employ the Llama-adapter~\cite{zhang2023llamaadapter} with an adaptation embedding length of 50 for efficient finetuning.
We utilize CLIP-ViT-L as the multimodal encoder. We employ a simple linear layer to project the CLIP feature into the LLM feature space.
The CLIP model and LLM backbone are frozen. The projection layer and additional Llama-adapter are trainable.
For text-only pre-alignment, 
we uniformly sample the {\Tideos} into 10 frames. 
We train the model on a mixture of tasks comprising {\Tideo} summarization, {\Tideo} QA, multi-choice {\Tideo} QA with the ratio of 1:1:2. 
TOPA-Llama2-7B and TOPA-Llama3-8B are trained on four 40G-A100 GPUs in one day. TOPA-Llama2-13B is trained in two days.
For zero-shot inference, we construct a memory for cross-modal projection, consisting of 2M CLIP text features sampled from the frame captions in the TextVid dataset.
We include the training details in Table~\ref{tab:trainingdetail}. 
The actual learning rate is calculated by $\texttt{base lr}*\texttt{Effective Batchsize}/256$.
\begin{table}[ht]
\small
\caption{Training hyper-parameters. 
}
\label{tab:trainingdetail}
\centering
\setlength{\tabcolsep}{0.5mm}
\begin{tabular}{l|c|c|c|c|c|c}
\tline
&\multirow{2}{*}{\textbf{Model}}&\multirow{2}{*}{\textbf{Training Dataset}}&\multirow{2}{*}{\textbf{Epoch}}&{\textbf{Effective Batchsize}}& \textbf{base}& \multirow{2}{*}{\textbf{Optimizer}} \\
&&&&(bs, \#GPUs, grad accu)&\textbf{lr}&\\
\hline
\multirow{3}{*}{Pre-training}&TOPA-LLama2-7B&\multirow{3}{*}{TextVid 721K}&\multirow{3}{*}{20}&18x4x4&5e-3&AdamW\\
&TOPA-LLama2-13B&&&4x4x8&8e-3&weight decay 0.1\\
&TOPA-LLama3-8B&&&14x4x8&5e-3&warm up 1 epoch\\
\hline

\multirow{9}{*}{Fine-tuning}&\multirow{3}{*}{TOPA-LLama2-7B}&NextQA&\multirow{9}{*}{5}&\multirow{3}{*}{20x4x4}&5e-3&\\
&&STAR&5&&1e-2&\\
&&TVQA&&&5e-3&\\
\cline{2-6}
&\multirow{3}{*}{TOPA-LLama2-13B}&NextQA&&\multirow{3}{*}{6x4x16}&2e-3&AdamW~\cite{adamw}\\
&&STAR&&&2e-3&weight decay 0.1\\
&&TVQA&&&5e-3&warm up 1 epochs\\
\cline{2-6}
&\multirow{3}{*}{TOPA-LLama3-8B}&NextQA&&\multirow{3}{*}{20x4x4}&1e-2&\\
&&STAR&5&&1e-2&\\
&&TVQA&&&5e-3&\\

\hline
\multirow{9}{*}{Baseline}&\multirow{3}{*}{LLama2-7B}&NextQA&\multirow{9}{*}{10}&\multirow{3}{*}{20x4x4}&1e-2&\\
&&STAR&10&&2e-2&\\
&&TVQA&&&2e-2&\\
\cline{2-6}
&\multirow{3}{*}{LLama2-13B}&NextQA&&\multirow{3}{*}{6x4x16}&1e-2&AdamW\\
&&STAR&10&&2e-2&weight decay 0.1\\
&&TVQA&&&2e-2&warm up 2 epochs\\
\cline{2-6}
&\multirow{3}{*}{LLama3-8B}&NextQA&&\multirow{3}{*}{20x4x4}&2e-2&\\
&&STAR&10&&2e-2&\\
&&TVQA&&&1e-2&\\
\tline

\end{tabular}
\end{table}

\section{Prompts} \label{app:prompts}
\subsection{Text-only Training Prompts.} \label{app:text-only task prompts}
We use the following prompts for Text-only training. The prompts are partially based on \cite{LLamavqa}.

\colorbox{gray!10}{%
\begin{minipage}{0.9\textwidth}
\Tideo~Multi-choice QA: \\
\textcolor{navy}{Instruction: Choose the correct answer based on the video and question. \\
Video: \{$\mathbf{f}^t_1$,...,$\mathbf{f}^t_n$\}.\\
~Question:~\{Question\}.\\
Choices: \\
(A):~\{Option A\}.~(B):~\{Option B\}.~(C):~\{Option C\}.~(D):~\{Option D\}.~(E):~\{Option E\}. \\
Answer:~The correct choice is~} \textcolor{maroon}{\{Correct Choice\}}. \\

\Tideo~QA: \\
\textcolor{navy}{Instruction: Predict the answer based on the video and question. \\
Video: \{$\mathbf{f}^t_1$,...,$\mathbf{f}^t_n$\}.\\
Question:~\{Question\}.  \\
Answer:} \textcolor{maroon}{\{Answer\}}. \\

\Tideo~Description: \\
\textcolor{navy}{Instruction: Generate a dense description for the video.\\
Video: \{$\mathbf{f}^t_1$,...,$\mathbf{f}^t_n$\}. \\
Description:} \textcolor{maroon}{\{Tideo Description\}}.
\end{minipage}
}


\subsection{Prompt for Gemini Blind Evaluation}
In Table~\ref{tab:blindVQA}, we use the following prompt for the blind evaluation of Gemini-Pro-1.0 on EgoSchema. The prompt is based on \cite{MC:MemoryConsolidation}.
\begin{figure*}[ht]
    \centering
    \includegraphics[width=0.8\textwidth]{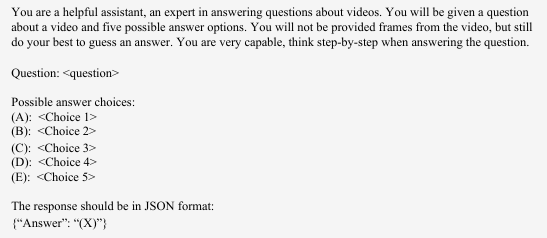}
    \vspace{-0.5em}
    \caption{The multi-choice QA prompts used for the blind evaluation of Gemini-1.0-Pro.}
    \label{fig:blind_prompt}
\end{figure*}

\subsection{Prompts for Data Generation}
\begin{figure*}[hb]
    \centering
    \includegraphics[width=0.81\textwidth]{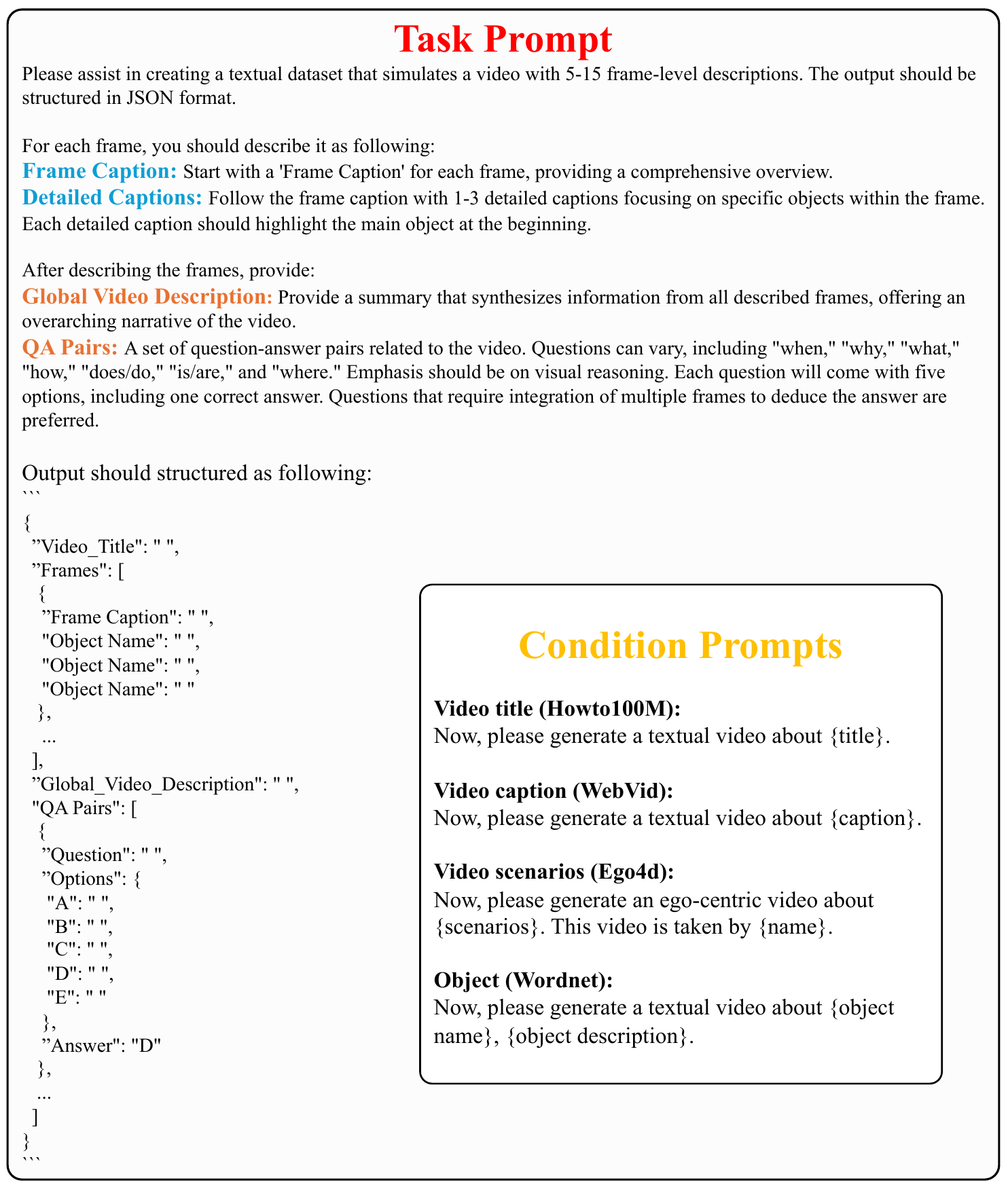}
    \vspace{-0.3cm}
    \caption{The prompts used in TextVid generation.}
    \label{fig:prompts}
    \vspace{-0.1cm}
\end{figure*}

The prompts for TextVid generation are shown in Figure~\ref{fig:prompts}. 

\clearpage
\section{License} \label{sec:license}
The code, model and proposed dataset will be publicly accessible. We use standard licenses from the community. We include the following licenses for the codes, datasets and models we used in this paper. 
\begin{enumerate}  
\item {\bf Benchmarks}
\item[] {NExT-QA~\cite{nextqa}}: \href{https://github.com/doc-doc/NExT-QA/blob/main/LICENSE}{MIT}

\item[] {STAR~\cite{star}}: \href{https://github.com/alexdobin/STAR/blob/master/LICENSE}{Apache}

\item[] {TVQA~\cite{tvqa}}: \href{https://github.com/jayleicn/TVQA/blob/master/LICENSE}{MIT}

\item[] {EgoSchema~\cite{EgoSchema}}:
\href{https://github.com/facebookresearch/Ego4d/blob/main/LICENSE}{MIT}

\item[] {MVBench~\cite{li2023mvbench}}:
\href{https://github.com/OpenGVLab/Ask-Anything/blob/main/LICENSE}{MIT}

\item {\bf Codes}

\item[] {LLama-adapter~\cite{zhang2023llamaadapter}} \href{https://github.com/OpenGVLab/LLaMA-Adapter/blob/main/LICENSE}{GNU General Public License v3.0}

\item[]{Flipped-VQA
~\cite{LLamavqa}}: \href{https://github.com/mlvlab/Flipped-VQA/blob/master/LICENSE}{MIT}

\item {\bf Models}

\item[]{CLIP~\cite{clip}}:
\href{https://github.com/openai/CLIP/blob/main/LICENSE}{MIT}

\item[]{LLama2~\cite{llama2}}:
\href{https://github.com/meta-llama/llama/blob/main/LICENSE}{Llama 2 Community License Agreement}

\item[]{LLama3}: \href{https://github.com/meta-llama/llama3/blob/main/LICENSE}{Meta Llama 3 Community License Agreement}

\item[]{Gimini-API~\cite{gemini}}:
\href{https://ai.google.dev/gemini-api/terms}{Gemini API Additional Terms of Service.}
\end{enumerate}

\clearpage
\section{Examples of TextVid} \label{sec:examples}
\begin{figure*}[ht]
    \centering
    \includegraphics[width=1.0\textwidth]{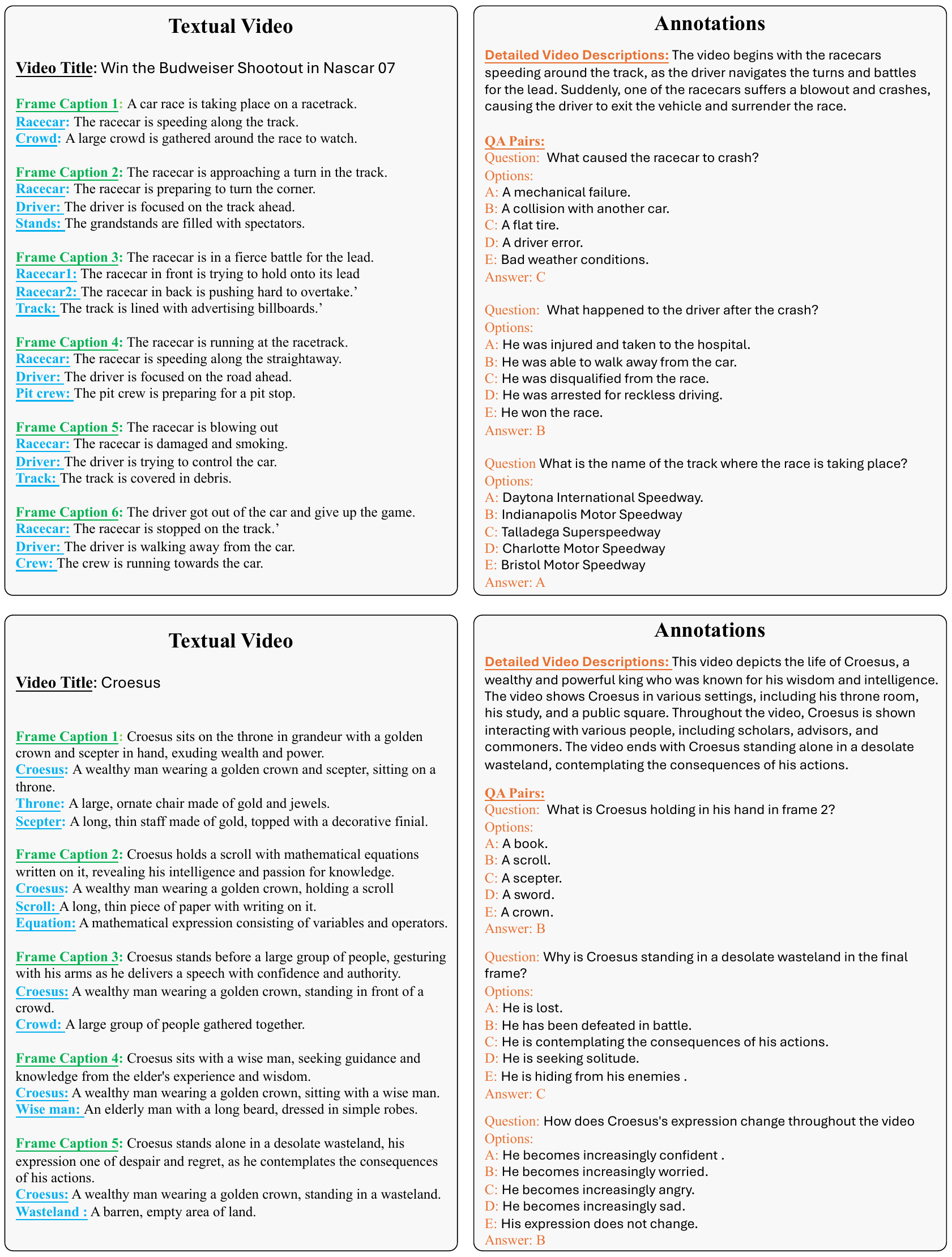}
\end{figure*}

\begin{figure*}[ht]
    \centering
    \includegraphics[width=1.0\textwidth]{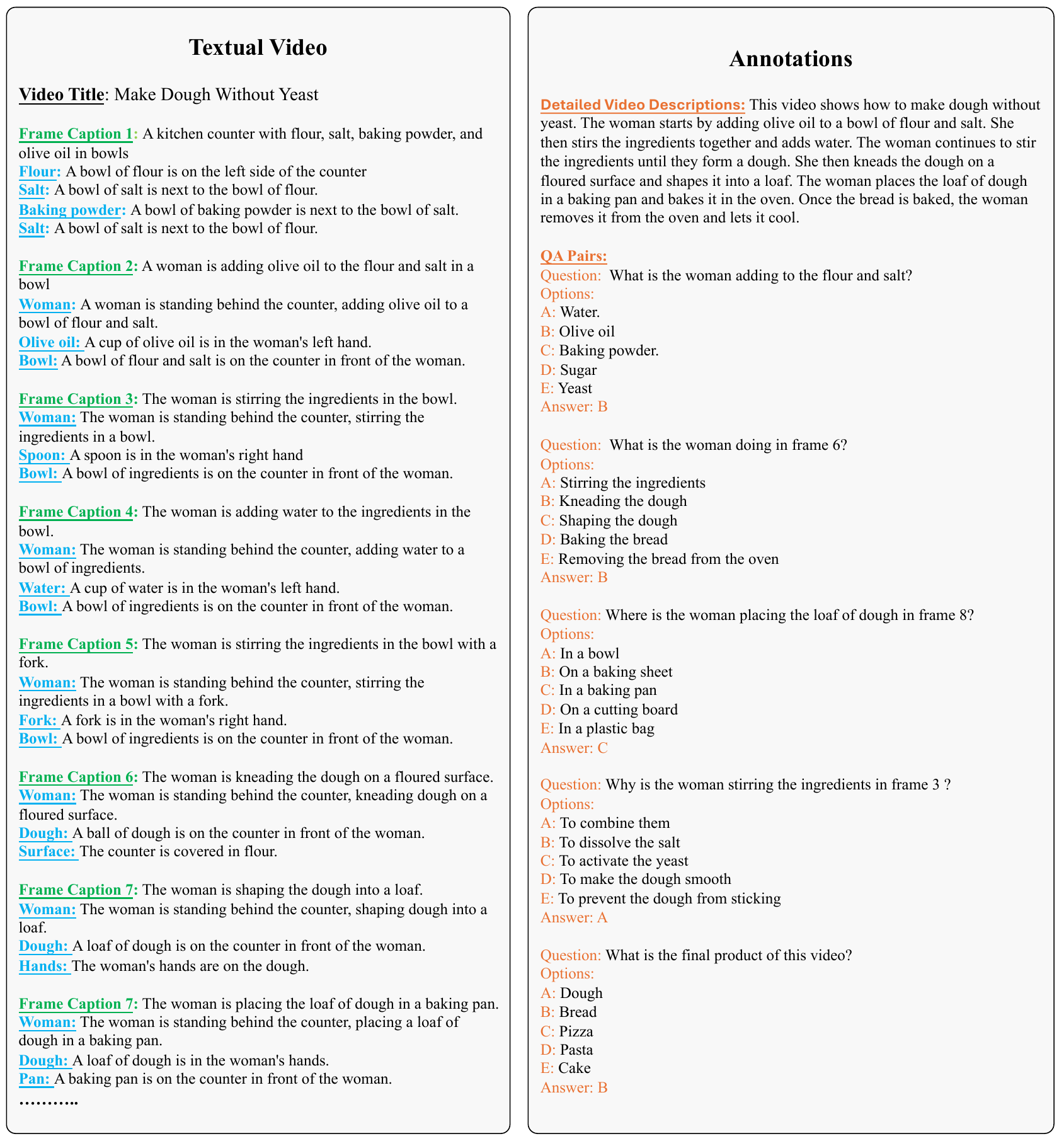}
\end{figure*}

\begin{figure*}[ht]
    \centering
    \includegraphics[width=1.0\textwidth]{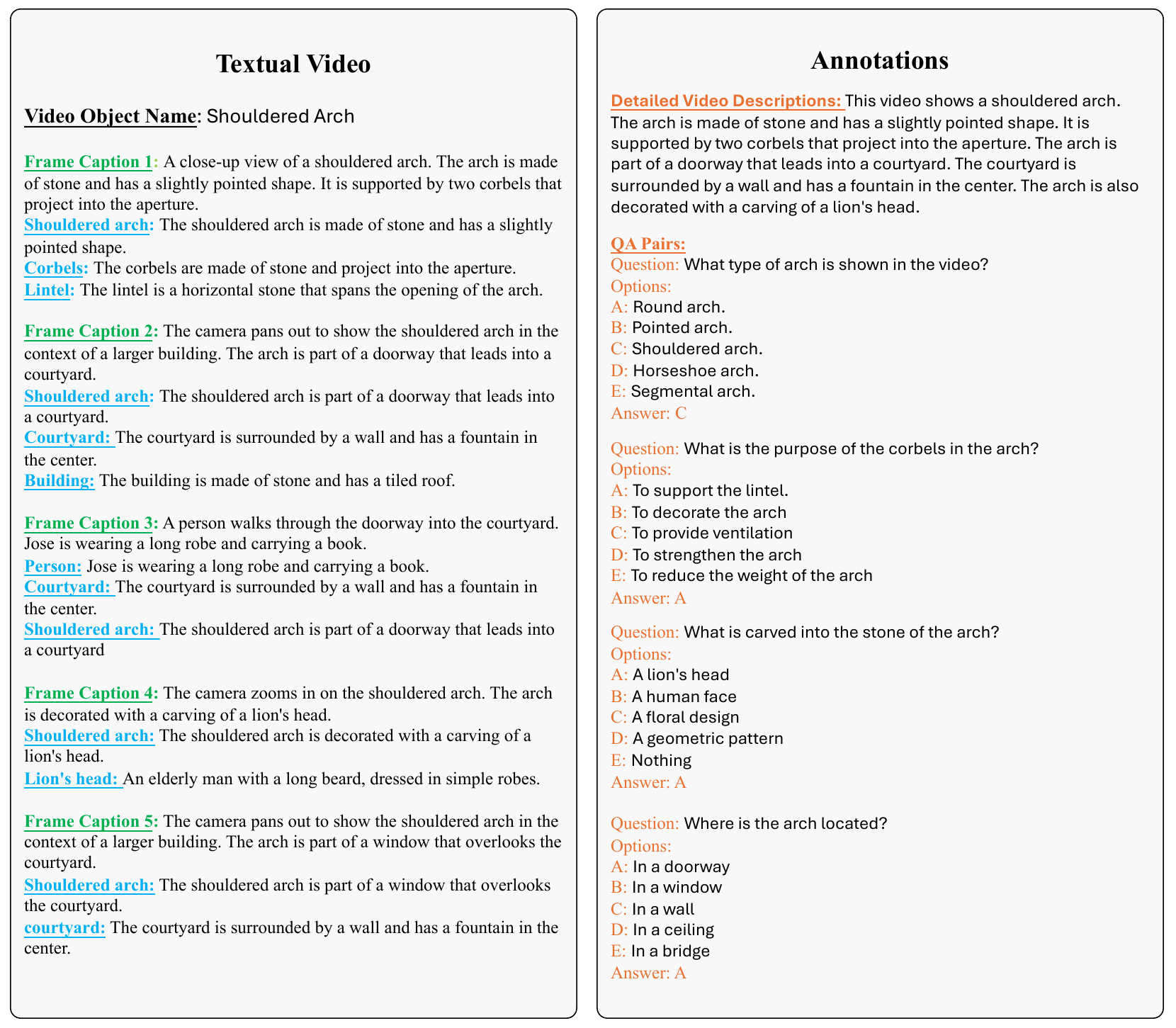}
\end{figure*}

\begin{figure*}[ht]
    \centering
    \includegraphics[width=1.0\textwidth]{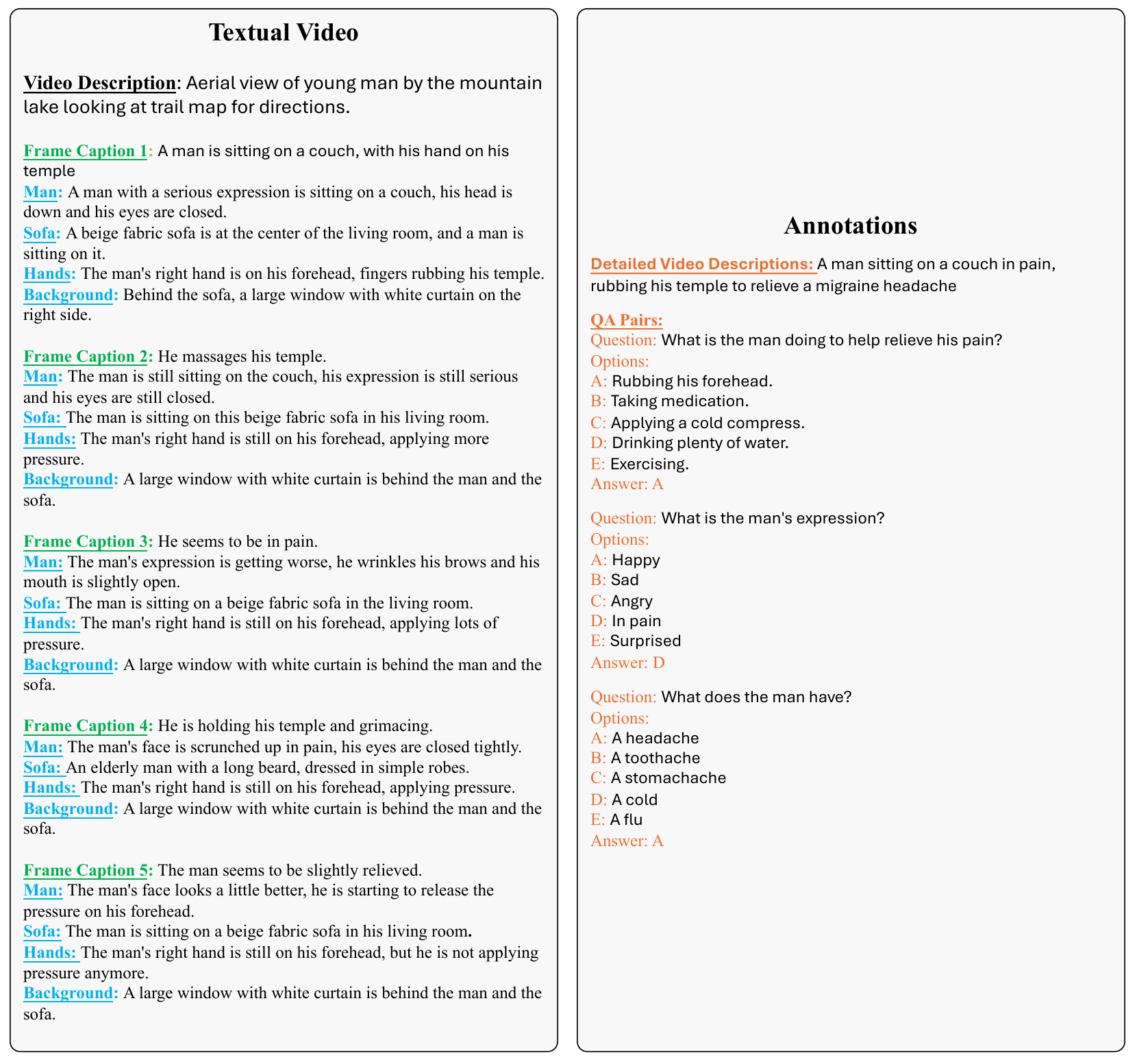}
\end{figure*}

\begin{figure*}[ht]
    \centering
    \includegraphics[width=1.0\textwidth]{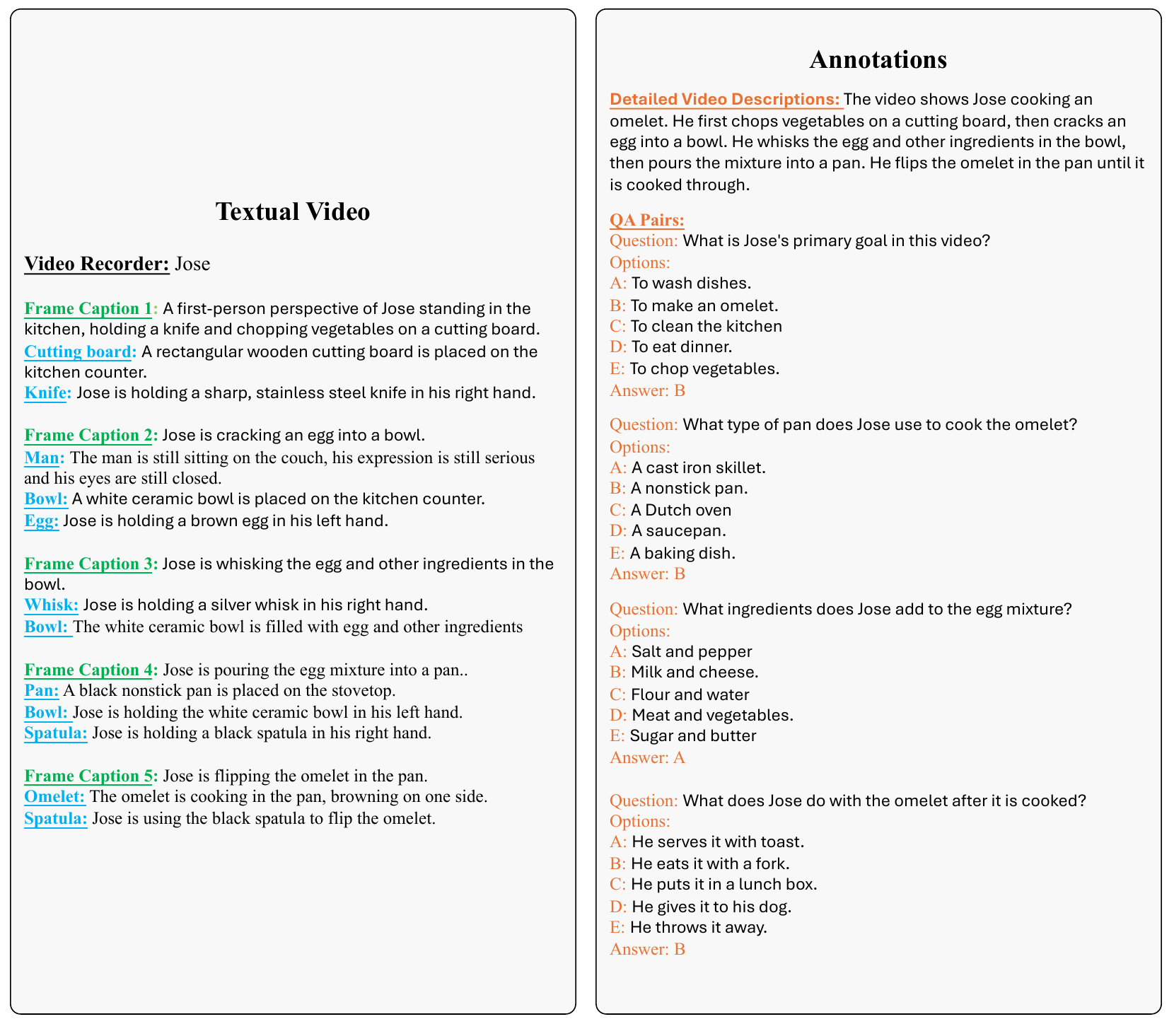}
\end{figure*}

\clearpage



\newpage

\newpage

\section*{NeurIPS Paper Checklist}

\begin{enumerate}

\item {\bf Claims}
    \item[] Question: Do the main claims made in the abstract and introduction accurately reflect the paper's contributions and scope?
    \item[] Answer: \answerYes{} 
    \item[] Justification: This paper aim to extend Large Language Models for video understanding via Text-Only Pre-Alignment, which does not need real videos for pre-training. Extensive experiments on video understanding benchmarks demonstrate the effectiveness of TOPA.
    \item[] Guidelines:
    \begin{itemize}
        \item The answer NA means that the abstract and introduction do not include the claims made in the paper.
        \item The abstract and/or introduction should clearly state the claims made, including the contributions made in the paper and important assumptions and limitations. A No or NA answer to this question will not be perceived well by the reviewers. 
        \item The claims made should match theoretical and experimental results, and reflect how much the results can be expected to generalize to other settings. 
        \item It is fine to include aspirational goals as motivation as long as it is clear that these goals are not attained by the paper. 
    \end{itemize}

\item {\bf Limitations}
    \item[] Question: Does the paper discuss the limitations of the work performed by the authors?
    \item[] Answer: \answerYes{} 
    \item[] Justification: The limitations of TOPA are detailed in Appendix~\ref{sec:limitation}.
    \item[] Guidelines:
    \begin{itemize}
        \item The answer NA means that the paper has no limitation while the answer No means that the paper has limitations, but those are not discussed in the paper. 
        \item The authors are encouraged to create a separate "Limitations" section in their paper.
        \item The paper should point out any strong assumptions and how robust the results are to violations of these assumptions (e.g., independence assumptions, noiseless settings, model well-specification, asymptotic approximations only holding locally). The authors should reflect on how these assumptions might be violated in practice and what the implications would be.
        \item The authors should reflect on the scope of the claims made, e.g., if the approach was only tested on a few datasets or with a few runs. In general, empirical results often depend on implicit assumptions, which should be articulated.
        \item The authors should reflect on the factors that influence the performance of the approach. For example, a facial recognition algorithm may perform poorly when image resolution is low or images are taken in low lighting. Or a speech-to-text system might not be used reliably to provide closed captions for online lectures because it fails to handle technical jargon.
        \item The authors should discuss the computational efficiency of the proposed algorithms and how they scale with dataset size.
        \item If applicable, the authors should discuss possible limitations of their approach to address problems of privacy and fairness.
        \item While the authors might fear that complete honesty about limitations might be used by reviewers as grounds for rejection, a worse outcome might be that reviewers discover limitations that aren't acknowledged in the paper. The authors should use their best judgment and recognize that individual actions in favor of transparency play an important role in developing norms that preserve the integrity of the community. Reviewers will be specifically instructed to not penalize honesty concerning limitations.
    \end{itemize}

\item {\bf Theory Assumptions and Proofs}
    \item[] Question: For each theoretical result, does the paper provide the full set of assumptions and a complete (and correct) proof?
    \item[] Answer: \answerNA{} 
    \item[] Justification: The paper does not include theoretical results. 
    \item[] Guidelines:
    \begin{itemize}
        \item The answer NA means that the paper does not include theoretical results. 
        \item All the theorems, formulas, and proofs in the paper should be numbered and cross-referenced.
        \item All assumptions should be clearly stated or referenced in the statement of any theorems.
        \item The proofs can either appear in the main paper or the supplemental material, but if they appear in the supplemental material, the authors are encouraged to provide a short proof sketch to provide intuition. 
        \item Inversely, any informal proof provided in the core of the paper should be complemented by formal proofs provided in appendix or supplemental material.
        \item Theorems and Lemmas that the proof relies upon should be properly referenced. 
    \end{itemize}

    \item {\bf Experimental Result Reproducibility}
    \item[] Question: Does the paper fully disclose all the information needed to reproduce the main experimental results of the paper to the extent that it affects the main claims and/or conclusions of the paper (regardless of whether the code and data are provided or not)?
    \item[] Answer: \answerYes{} 
    \item[] Justification: The training details of TOPA are included in Appendix~\ref{app:training_details}. The dataset generation pipeline is detailed in Appendix~\ref{app:dataset}.
    \item[] Guidelines:
    \begin{itemize}
        \item The answer NA means that the paper does not include experiments.
        \item If the paper includes experiments, a No answer to this question will not be perceived well by the reviewers: Making the paper reproducible is important, regardless of whether the code and data are provided or not.
        \item If the contribution is a dataset and/or model, the authors should describe the steps taken to make their results reproducible or verifiable. 
        \item Depending on the contribution, reproducibility can be accomplished in various ways. For example, if the contribution is a novel architecture, describing the architecture fully might suffice, or if the contribution is a specific model and empirical evaluation, it may be necessary to either make it possible for others to replicate the model with the same dataset, or provide access to the model. In general. releasing code and data is often one good way to accomplish this, but reproducibility can also be provided via detailed instructions for how to replicate the results, access to a hosted model (e.g., in the case of a large language model), releasing of a model checkpoint, or other means that are appropriate to the research performed.
        \item While NeurIPS does not require releasing code, the conference does require all submissions to provide some reasonable avenue for reproducibility, which may depend on the nature of the contribution. For example
        \begin{enumerate}
            \item If the contribution is primarily a new algorithm, the paper should make it clear how to reproduce that algorithm.
            \item If the contribution is primarily a new model architecture, the paper should describe the architecture clearly and fully.
            \item If the contribution is a new model (e.g., a large language model), then there should either be a way to access this model for reproducing the results or a way to reproduce the model (e.g., with an open-source dataset or instructions for how to construct the dataset).
            \item We recognize that reproducibility may be tricky in some cases, in which case authors are welcome to describe the particular way they provide for reproducibility. In the case of closed-source models, it may be that access to the model is limited in some way (e.g., to registered users), but it should be possible for other researchers to have some path to reproducing or verifying the results.
        \end{enumerate}
    \end{itemize}

\item {\bf Open access to data and code}
    \item[] Question: Does the paper provide open access to the data and code, with sufficient instructions to faithfully reproduce the main experimental results, as described in supplemental material?
    \item[] Answer: \answerNo{} 
    \item[] Justification: The code and dataset is publicly accessible (\url{https://github.com/dhg-wei/TOPA}).
    \item[] Guidelines:
    \begin{itemize}
        \item The answer NA means that paper does not include experiments requiring code.
        \item Please see the NeurIPS code and data submission guidelines (\url{https://nips.cc/public/guides/CodeSubmissionPolicy}) for more details.
        \item While we encourage the release of code and data, we understand that this might not be possible, so “No” is an acceptable answer. Papers cannot be rejected simply for not including code, unless this is central to the contribution (e.g., for a new open-source benchmark).
        \item The instructions should contain the exact command and environment needed to run to reproduce the results. See the NeurIPS code and data submission guidelines (\url{https://nips.cc/public/guides/CodeSubmissionPolicy}) for more details.
        \item The authors should provide instructions on data access and preparation, including how to access the raw data, preprocessed data, intermediate data, and generated data, etc.
        \item The authors should provide scripts to reproduce all experimental results for the new proposed method and baselines. If only a subset of experiments are reproducible, they should state which ones are omitted from the script and why.
        \item At submission time, to preserve anonymity, the authors should release anonymized versions (if applicable).
        \item Providing as much information as possible in supplemental material (appended to the paper) is recommended, but including URLs to data and code is permitted.
    \end{itemize}

\item {\bf Experimental Setting/Details}
    \item[] Question: Does the paper specify all the training and test details (e.g., data splits, hyperparameters, how they were chosen, type of optimizer, etc.) necessary to understand the results?
    \item[] Answer: \answerYes{} 
    \item[] Justification: The details of training and evaluation are included in Appendix~\ref{app:training_details}.
    \item[] Guidelines:
    \begin{itemize}
        \item The answer NA means that the paper does not include experiments.
        \item The experimental setting should be presented in the core of the paper to a level of detail that is necessary to appreciate the results and make sense of them.
        \item The full details can be provided either with the code, in appendix, or as supplemental material.
    \end{itemize}

\item {\bf Experiment Statistical Significance}
    \item[] Question: Does the paper report error bars suitably and correctly defined or other appropriate information about the statistical significance of the experiments?
    \item[] Answer: \answerNo{} 
    \item[] Justification: TOPA is a large-scale pre-training framework. Similar to previous related work~\cite{longvivit,sevila,li2023mvbench}, error bars are not reported because it would be computationally too expensive.
    \item[] Guidelines:
    \begin{itemize}
        \item The answer NA means that the paper does not include experiments.
        \item The authors should answer "Yes" if the results are accompanied by error bars, confidence intervals, or statistical significance tests, at least for the experiments that support the main claims of the paper.
        \item The factors of variability that the error bars are capturing should be clearly stated (for example, train/test split, initialization, random drawing of some parameter, or overall run with given experimental conditions).
        \item The method for calculating the error bars should be explained (closed form formula, call to a library function, bootstrap, etc.)
        \item The assumptions made should be given (e.g., Normally distributed errors).
        \item It should be clear whether the error bar is the standard deviation or the standard error of the mean.
        \item It is OK to report 1-sigma error bars, but one should state it. The authors should preferably report a 2-sigma error bar than state that they have a 96\% CI, if the hypothesis of Normality of errors is not verified.
        \item For asymmetric distributions, the authors should be careful not to show in tables or figures symmetric error bars that would yield results that are out of range (e.g. negative error rates).
        \item If error bars are reported in tables or plots, The authors should explain in the text how they were calculated and reference the corresponding figures or tables in the text.
    \end{itemize}

\item {\bf Experiments Compute Resources}
    \item[] Question: For each experiment, does the paper provide sufficient information on the computer resources (type of compute workers, memory, time of execution) needed to reproduce the experiments?
    \item[] Answer: \answerYes{} 
    \item[] Justification: The details of compute resources are included in Section~\ref{exp_details}.
    \item[] Guidelines:
    \begin{itemize}
        \item The answer NA means that the paper does not include experiments.
        \item The paper should indicate the type of compute workers CPU or GPU, internal cluster, or cloud provider, including relevant memory and storage.
        \item The paper should provide the amount of compute required for each of the individual experimental runs as well as estimate the total compute. 
        \item The paper should disclose whether the full research project required more compute than the experiments reported in the paper (e.g., preliminary or failed experiments that didn't make it into the paper). 
    \end{itemize}
    
\item {\bf Code Of Ethics}
    \item[] Question: Does the research conducted in the paper conform, in every respect, with the NeurIPS Code of Ethics \url{https://neurips.cc/public/EthicsGuidelines}?
    \item[] Answer: \answerYes{} 
    \item[] Justification: We have reviewed the Code of Ethics and it conforms with the Code of Ethics.
    \item[] Guidelines:
    \begin{itemize}
        \item The answer NA means that the authors have not reviewed the NeurIPS Code of Ethics.
        \item If the authors answer No, they should explain the special circumstances that require a deviation from the Code of Ethics.
        \item The authors should make sure to preserve anonymity (e.g., if there is a special consideration due to laws or regulations in their jurisdiction).
    \end{itemize}

\item {\bf Broader Impacts}
    \item[] Question: Does the paper discuss both potential positive societal impacts and negative societal impacts of the work performed?
    \item[] Answer: \answerYes{} 
    \item[] Justification: The details of broader impacts are included in Appendix~\ref{sec:boarder impact}.
    \item[] Guidelines:
    \begin{itemize}
        \item The answer NA means that there is no societal impact of the work performed.
        \item If the authors answer NA or No, they should explain why their work has no societal impact or why the paper does not address societal impact.
        \item Examples of negative societal impacts include potential malicious or unintended uses (e.g., disinformation, generating fake profiles, surveillance), fairness considerations (e.g., deployment of technologies that could make decisions that unfairly impact specific groups), privacy considerations, and security considerations.
        \item The conference expects that many papers will be foundational research and not tied to particular applications, let alone deployments. However, if there is a direct path to any negative applications, the authors should point it out. For example, it is legitimate to point out that an improvement in the quality of generative models could be used to generate deepfakes for disinformation. On the other hand, it is not needed to point out that a generic algorithm for optimizing neural networks could enable people to train models that generate Deepfakes faster.
        \item The authors should consider possible harms that could arise when the technology is being used as intended and functioning correctly, harms that could arise when the technology is being used as intended but gives incorrect results, and harms following from (intentional or unintentional) misuse of the technology.
        \item If there are negative societal impacts, the authors could also discuss possible mitigation strategies (e.g., gated release of models, providing defenses in addition to attacks, mechanisms for monitoring misuse, mechanisms to monitor how a system learns from feedback over time, improving the efficiency and accessibility of ML).
    \end{itemize}
    
\item {\bf Safeguards}
    \item[] Question: Does the paper describe safeguards that have been put in place for responsible release of data or models that have a high risk for misuse (e.g., pretrained language models, image generators, or scraped datasets)?
    \item[] Answer: \answerNA{}{} 
    \item[] Justification: Our model and data focus on video-language understanding, with minimal risk of misuse.
    \item[] Guidelines:
    \begin{itemize}
        \item The answer NA means that the paper poses no such risks.
        \item Released models that have a high risk for misuse or dual-use should be released with necessary safeguards to allow for controlled use of the model, for example by requiring that users adhere to usage guidelines or restrictions to access the model or implementing safety filters. 
        \item Datasets that have been scraped from the Internet could pose safety risks. The authors should describe how they avoided releasing unsafe images.
        \item We recognize that providing effective safeguards is challenging, and many papers do not require this, but we encourage authors to take this into account and make a best faith effort.
    \end{itemize}

\item {\bf Licenses for existing assets}
    \item[] Question: Are the creators or original owners of assets (e.g., code, data, models), used in the paper, properly credited and are the license and terms of use explicitly mentioned and properly respected?
    \item[] Answer: \answerYes{} 
    \item[] Justification: The licenses are mentioned in Appendix~\ref{sec:license}.
    \item[] Guidelines:
    \begin{itemize}
        \item The answer NA means that the paper does not use existing assets.
        \item The authors should cite the original paper that produced the code package or dataset.
        \item The authors should state which version of the asset is used and, if possible, include a URL.
        \item The name of the license (e.g., CC-BY 4.0) should be included for each asset.
        \item For scraped data from a particular source (e.g., website), the copyright and terms of service of that source should be provided.
        \item If assets are released, the license, copyright information, and terms of use in the package should be provided. For popular datasets, \url{paperswithcode.com/datasets} has curated licenses for some datasets. Their licensing guide can help determine the license of a dataset.
        \item For existing datasets that are re-packaged, both the original license and the license of the derived asset (if it has changed) should be provided.
        \item If this information is not available online, the authors are encouraged to reach out to the asset's creators.
    \end{itemize}

\item {\bf New Assets}
    \item[] Question: Are new assets introduced in the paper well documented and is the documentation provided alongside the assets?
    \item[] Answer: \answerYes{} 
    \item[] Justification: The details of the new dataset and model are detailed in this paper.
    \item[] Guidelines:
    \begin{itemize}
        \item The answer NA means that the paper does not release new assets.
        \item Researchers should communicate the details of the dataset/code/model as part of their submissions via structured templates. This includes details about training, license, limitations, etc. 
        \item The paper should discuss whether and how consent was obtained from people whose asset is used.
        \item At submission time, remember to anonymize your assets (if applicable). You can either create an anonymized URL or include an anonymized zip file.
    \end{itemize}

\item {\bf Crowdsourcing and Research with Human Subjects}
    \item[] Question: For crowdsourcing experiments and research with human subjects, does the paper include the full text of instructions given to participants and screenshots, if applicable, as well as details about compensation (if any)? 
    \item[] Answer: \answerNA{} 
    \item[] Justification: The paper does not involve crowdsourcing nor research with human subjects.
    \item[] Guidelines:
    \begin{itemize}
        \item The answer NA means that the paper does not involve crowdsourcing nor research with human subjects.
        \item Including this information in the supplemental material is fine, but if the main contribution of the paper involves human subjects, then as much detail as possible should be included in the main paper. 
        \item According to the NeurIPS Code of Ethics, workers involved in data collection, curation, or other labor should be paid at least the minimum wage in the country of the data collector. 
    \end{itemize}

\item {\bf Institutional Review Board (IRB) Approvals or Equivalent for Research with Human Subjects}
    \item[] Question: Does the paper describe potential risks incurred by study participants, whether such risks were disclosed to the subjects, and whether Institutional Review Board (IRB) approvals (or an equivalent approval/review based on the requirements of your country or institution) were obtained?
    \item[] Answer: \answerNA{} 
    \item[] Justification: The paper does not involve crowdsourcing nor research with human subjects.
    \item[] Guidelines:
    \begin{itemize}
        \item The answer NA means that the paper does not involve crowdsourcing nor research with human subjects.
        \item Depending on the country in which research is conducted, IRB approval (or equivalent) may be required for any human subjects research. If you obtained IRB approval, you should clearly state this in the paper. 
        \item We recognize that the procedures for this may vary significantly between institutions and locations, and we expect authors to adhere to the NeurIPS Code of Ethics and the guidelines for their institution. 
        \item For initial submissions, do not include any information that would break anonymity (if applicable), such as the institution conducting the review.
    \end{itemize}

\end{enumerate}

\end{document}